\documentclass[10pt,twocolumn,letterpaper]{article}

\usepackage[pagenumbers]{cvpr} 

\usepackage{graphicx}
\usepackage{amsmath}
\usepackage{amssymb}
\usepackage{booktabs}
\usepackage{pifont}

\usepackage{booktabs}

\usepackage{capt-of}
\usepackage[utf8]{inputenc} 
\usepackage[T1]{fontenc}    

\usepackage{url}            
\usepackage{amsfonts}       
\usepackage{nicefrac}       
\usepackage{microtype}      
\usepackage{xcolor}         
\usepackage{multirow}

\usepackage{makecell}
\usepackage{colortbl}

\usepackage{siunitx}
\usepackage{bbding}

\usepackage{paralist}

\usepackage[pagebackref=true,breaklinks=true,letterpaper=true,colorlinks,citecolor=blue,linkcolor=red,bookmarks=false]{hyperref}

\usepackage[capitalize]{cleveref}
\crefname{section}{Sec.}{Secs.}
\Crefname{section}{Section}{Sections}
\Crefname{table}{Table}{Tables}
\crefname{table}{Tab.}{Tabs.}

\def\cvprPaperID{3818} 
\def\confName{CVPR}
\def\confYear{2023}

\begin{document}

\title{Efficient Frequency Domain-based Transformers \\for High-Quality Image Deblurring}

\author{Lingshun Kong$^{1}$\quad Jiangxin Dong$^{1}$\quad Mingqiang Li$^{2}$ \quad Jianjun Ge$^{2}$ \quad Jinshan Pan$^{1}$ \\
$^{1}$Nanjing University of Science and Technology \quad $^{2}$CETC\\
\\
}
\maketitle

\begin{abstract}
We present an effective and efficient method that explores the properties of Transformers in the frequency domain for high-quality image deblurring.
Our method is motivated by the convolution theorem that the correlation or convolution of two signals in the spatial domain is equivalent to an element-wise product of them in the frequency domain.
This inspires us to develop an efficient frequency domain-based self-attention solver (FSAS) to estimate the scaled dot-product attention by an element-wise product operation instead of the matrix multiplication in the spatial domain.
In addition, we note that simply using the naive feed-forward network (FFN) in Transformers does not generate good deblurred results. To overcome this problem, we propose a simple yet effective discriminative frequency domain-based FFN (DFFN), where we introduce a gated mechanism in the FFN based on the Joint Photographic Experts Group (JPEG) compression algorithm to discriminatively determine which low- and high-frequency information of the features should be preserved for latent clear image restoration.
We formulate the proposed FSAS and DFFN into an asymmetrical network based on an encoder and decoder architecture, where the FSAS is only used in the decoder module for better image deblurring.
Experimental results show that the proposed method performs favorably against the state-of-the-art approaches. Code will be available at \url{https://github.com/kkkls/FFTformer}.

\end{abstract}

\begin{figure}[!t]
    \centering
    \vspace{0mm}
 \includegraphics[width=0.48\textwidth]{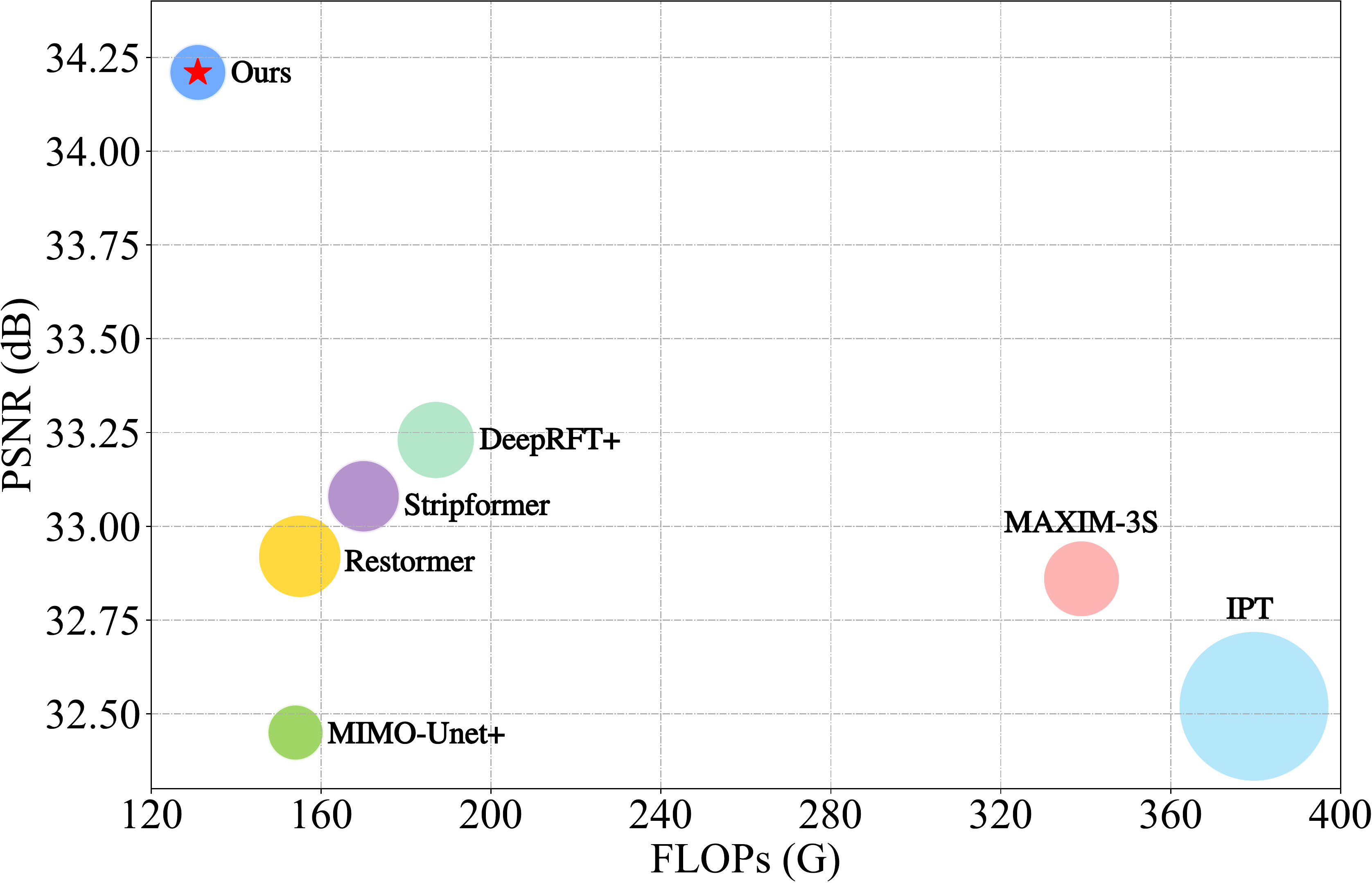}
 \vspace{-6mm}
 \caption{Comparisons of the proposed method and state-of-the-art ones on the GoPro dataset \cite{GoPro} in terms of accuracy, floating point operations (FLOPs), and network parameters. The circle size indicates the number of the network parameter. }
 \label{fig:Params}
 \vspace{-5mm}
\end{figure}

\vspace{-2mm}
\section{Introduction}
\label{sec:intro}
Image deblurring aims to restore high-quality images from blurred ones. This problem has achieved significant progress due to the development of various effective deep models with large-scale training datasets.

Most state-of-the-art methods for image deblurring are mainly based on deep convolutional neural networks (CNNs). The main success of these methods is due to developing kinds of network architectural designs, for example, the multi-scale~\cite{GoPro,SRN,MIMO} or multi-stage~\cite{DMPHN,MPRNet} network architectures, generative adversarial learning~\cite{DeblurGAN,DeblurGANv2}, physics model inspired network structures~\cite{svrnn,physicgan}, and so on.
As the basic operation in these networks, the convolution operation is a spatially-invariant local operation, which does not model the spatially variant properties of the image contents. Most of them use larger and deeper models to remedy the limitation of the convolution. However, simply increasing the capacity of deep models does not always lead to better performance as shown in~\cite{svrnn,physicgan}.

Different from the convolution operation that models the local connectivity, Transformers are able to model the global contexts by computing the correlations of one token to all other tokens. They have been shown to be an effective approach in lots of high-level vision tasks and also have great potential to be the alternatives of deep CNN models.
In image deblurring, the methods based on Transformers~\cite{Restormer,Uformer} also achieve better performance than the CNN-based methods.
However, the computation of the scaled dot-product attention in Transformers leads to quadratic space and time complexity in terms of the number of tokens.
Although using smaller and fewer tokens can reduce the space and time complexity, such strategy cannot model the long-range information of features well and usually leads to significant artifacts when handling high-resolution images, which thus limits the performance improvement.

To alleviate this problem, most approaches use the downsampling strategy to reduce the spatial resolution of features~\cite{PyramidVIT}. However, reducing the spatial resolution of features will cause information loss and thus affect the image deblurring.
Several methods reduce the computational cost by computing the scaled dot-product attention in terms of the number of features~\cite{Restormer,cotransformer}.
Although the computational cost is reduced, the spatial information is well not explored, which may affect the deblurring performance.
%

In this paper, we develop an effective and efficient method that explores the properties of Transformers for high-quality image deblurring.
We note that the scaled dot-product attention computation is actually to estimate the correlation of one token from the query and all the tokens from the key. This process can be achieved by a convolution operation when rearranging the permutations of tokens.
Based on this observation and the convolution theorem that the convolution in the spatial domain equals a point-wise multiplication in the frequency domain,
we develop an efficient frequency domain-based self-attention solver (FSAS) to estimate the scaled dot-product attention by an element-wise product operation instead of the matrix multiplication.
Therefore, the space and time complexity can be reduced to $O(N)$ $O(N\log N)$ for each feature channel, where $N$ is the number of the pixels.

In addition, we note that simply using the feed-forward network (FFN) by~\cite{Restormer} does not generate good deblurred results.
To generate better features for latent clear image restoration, we develop a simple yet effective discriminative frequency domain-based FFN (DFFN).
Our DFFN is motivated by the Joint Photographic Experts Group (JPEG) compression algorithm. It introduces a gated mechanism in the FFN to discriminatively determine which low- and high-frequency information should be preserved for latent clear image restoration.
%

We formulate the proposed FSAS and DFFN into an end-to-end trainable network based on an encoder and decoder architecture to solve image deblurring.
However, we find that as features of shallow layers usually contain blur effects, applying the scaled dot-product attention to shallow features does not effectively explore global clear contents.
As the features from deep layers are usually clearer than those from shallow layers, we develop an asymmetric network architecture, where the FSAS is only used in the decoder module for better image deblurring.
We analyze that the exploring properties of Transformers in the frequency domain is able to facilitate blur removal. Experimental results demonstrate that the proposed method generates favorable results against state-of-the-art methods in terms of accuracy and efficiency (Figure~\ref{fig:Params}).

The main contributions of this work are summarized as follows:
\begin{compactitem}
    \item We develop an efficient frequency domain-based self-attention solver to estimate the scaled dot-product attention. Our analysis demonstrates that using the frequency domain-based solver reduces the space and time complexity and is much more effective and efficient.
    \item We propose a simple yet effective discriminative frequency domain-based FFN based on the JPEG compression algorithm to discriminatively determine which low and high-frequency information should be preserved for latent clear image restoration.
    \item We develop an asymmetric network architecture based on an encoder and decoder network, where the frequency domain-based self-attention solver is only used in the decoder module for better image deblurring.
    \item We analyze that the exploring properties of Transformers in the frequency domain is able to facilitate blur removal and show that our approach performs favorably against state-of-the-art methods.
\end{compactitem}

\section{Related Work}
\vspace{-1mm}
{\flushleft \textbf{Deep CNN-based Image deblurring methods.}}
In recent years, we have witnessed significant advances in image deblurring due to the development of different deep CNN models~\cite{GoPro,SRN,SSN,DMPHN,MPRNet,MIMO,NAFNet}.
In~\cite{GoPro}, Nah et al. propose a deep CNN based on a multi-scale framework to directly estimate clear images from blurred ones.
To better utilize the information of each scale in multi-scale framework, Tao et al.~\cite{SRN} develop an effective scale recurrent network.
Gao et al.~\cite{SSN} propose a selective network parameter sharing method to improve~\cite{GoPro,SRN}.

As using more scales does not improve the performance significantly, Zhang et al.~\cite{DMPHN} develop an effective network based on multi-patch strategy.
The deblurring process is achieved stage by stage.
To better explore the features from different stages, Zamir et al.~\cite{MPRNet} propose a cross-stage feature fusion for better performance.
In order to reduce the computational cost of the methods based on multi-scale framework, Cho et al.~\cite{MIMO} present a multi-input and multi-output network.
Chen et al.~\cite{NAFNet} analyze the baseline modules and simplify them for better image restoration.
As demonstrated in~\cite{Restormer}, the convolution operation is spatial invariant and does not effectively model the global contexts for image deblurring.
\vspace{-2mm}
{\flushleft \textbf{Transformers and their applications to image deblurring.}}
As the Transformer~\cite{Transformer} can model the global contexts and achieves significant progress in lots of high-level vision tasks (e.g., image classification~\cite{Swin}, object detection~\cite{Object_Detection,Object_Detection_2} and semantic segmentation~\cite{Segmentation,Segmentation_2}), it has been developed to solve image super-resolution~\cite{SwinIR}, image deblurring~\cite{Restormer,Stripformer} and image denoise~\cite{IPT,Uformer}.
To reduce the computational cost of Transformer, Zamir et al.~\cite{Restormer} propose an efficient Transformer model by computing the scaled dot-product attention in the feature depth domain.
This method can effectively explore information from different features along the channel dimension. However, the spatial information that is vital for image restoration is not fully explored.
Tsai et al.~\cite{Stripformer} simplify the calculation of self-attention by constructing intra and inter strip tokens to replace the global attention.
Wang et al.~\cite{Uformer} propose a Transformer based on a UNet which uses non-overlapping window-based self-attention for single image deblurring.
Although using the splitting strategy reduces the computational cost, the coarse splitting does not fully explore the information of each patch.
Moreover, the scaled dot-product attention in these methods usually needs the complex matrix multiplication whose the space and time complexity is quadratic.

Different from these methods, we develop an efficient Transformer-based method that explores the property of the frequency domain to avoid the complex matrix multiplication for the scaled dot-product attention.

\begin{figure*}[thbp]
    \centering
 \includegraphics[width=0.98\textwidth]{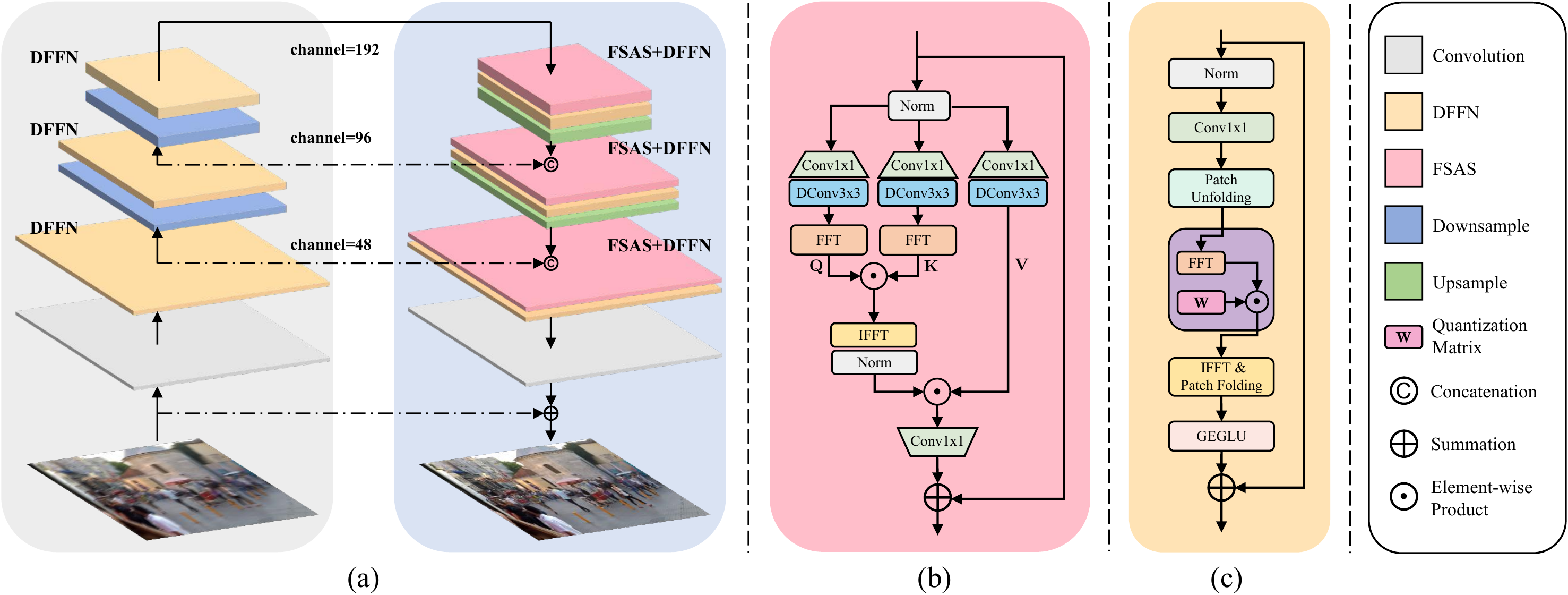}
 \vspace{-1mm}
 \caption{Network architectures. (a) The proposed asymmetric encoder-decoder network that only contains DFFN in the encoder module and both FSAS and DFFN in the decoder module for image deblurring. (b) The proposed FSAS module. (c) The proposed DFFN module.}
 \label{fig: Network}
 \vspace{-4mm}
\end{figure*}

\section{Proposed Method}

Our goal is to present an effective and efficient method to explore the properties of Transformers for high-quality image deblurring.
To this end, we first develop an efficient frequency domain-based self-attention solver to estimate the scaled dot-product attention. To refine the features estimated by the frequency domain-based solver, we further develop a discriminative frequency domain-based feed-forward network.
We formulate these above approaches into an end-to-end trainable network based on an encoder and decoder architecture to solve image deblurring, where the frequency domain-based self-attention solver for the estimation of the scaled dot-product attention is used in the decoder module for better feature representation.
Figure~\ref{fig: Network}(a) shows the overview of the proposed method. In the following, we present the details of each component.

\subsection{Frequency domain-based self-attention solver}
Given the input feature $X$ with a spatial resolution of $H\times W$ pixels and $C$ channels, existing vision Transformers usually first compute the features $F_q$, $F_k$, and $F_v$ by applying linear transformations ${W}_q$, ${W}_k$, and ${W}_v$ to $X$.
%
Then, they apply the unfolding function to the features $F_q$, $F_k$, and $F_v$ to extract image patches $\{q_i\}_{i= 1}^{N}$, $\{k_i\}_{i= 1}^{N}$, and $\{v_i\}_{i= 1}^{N}$, where $N$ denotes the number of extracted patches.
By applying a reshape operation to the extracted patches, the query $\mathbf{Q}$, key $\mathbf{K}$, and value $\mathbf{V}$ can be obtained by:
\begin{equation}
\label{eq: patch-divide}
\mathbf{Q} =  \mathcal{R}(\{q_i\}_{i= 1}^{N}), \
\mathbf{K} =  \mathcal{R}(\{k_i\}_{i= 1}^{N}), \
\mathbf{V} =  \mathcal{R}(\{v_i\}_{i= 1}^{N}), \\
\end{equation}
where $\mathcal{R}$ denotes the reshape function which ensures that $\{\mathbf{K}, \mathbf{Q}, \mathbf{V}\}\in R^{N\times (CH_pW_p)}$, $H_p$ and $W_p$ denote the height and width of extracted patches.
Based on the obtained query $\mathbf{Q}$, key $\mathbf{K}$, and value $\mathbf{V}$, the scaled dot-product attention is achieved by:
\begin{equation}
V_{att} = \mathrm{softmax}\left(\frac{\mathbf{QK}^{\top}}{\sqrt{CH_pW_p}}\right)V.
\label{eq: attention-transformer-ori}
\end{equation}
The attention map computation involves the matrix multiplication of $\mathbf{QK}^{\top}$ whose space complexity and time complexity are $O(N^2)$ and $O(N^2C)$.
It is not affordable if the image resolution and the number of the extracted patches are large.
Although using downsampling operation to reduce the image resolution or non-overlapping method to extract fewer patches will alleviate the problem,
these strategies would lead to information loss and limit the ability to model details within and across each patch~\cite{cotransformer}.

We note that each element of $\mathbf{QK}^{\top}$ is obtained by the inner product:
\begin{equation}
\left(\mathbf{QK}^{\top}\right)_{ij} = \left<\mathbf{q}_i,\mathbf{k}_{j}\right>,
\label{eq: correlation-operation-element-corr}
\end{equation}
where $\mathbf{q}_i$ and $\mathbf{k}_j$ are the vectorized forms of $i$-th and $j$-th patches from $F_q$ and $F_k$.
Based on~\eqref{eq: correlation-operation-element-corr}, if we apply reshape functions to $\mathbf{q}_i$ and all the patches $\mathbf{k}_{j}$, respectively, all the i-th column elements of $\mathbf{QK}^{\top}$ can be obtained by a convolution operation, i.e., $\widetilde{q}_i\otimes \widetilde{K}$, where $\widetilde{q}_i$ and $\widetilde{K}$ denote the reshaped results of $\mathbf{q}_i$ and $\mathbf{k}_{j}$; $\otimes$ denotes the convolution operation.

According to the convolution theorem, the correlation or convolution of two signals in the spatial domain is equivalent to an element-wise product of them in the frequency domain.
Therefore, \textit{a natural question is that can we efficiently estimate the attention map by an element-wise product operation in a frequency domain instead of computing the matrix multiplication of $\mathbf{QK}^{\top}$ in the spatial domain? }

To this end, we develop an effective frequency domain-based self-attention solver.
Specifically, we first obtain $F_q$, $F_k$, and $F_v$ by a $1\times 1$ point-wise convolution and $3\times 3$ depth-wise convolution. Then, we apply the fast Fourier transform (FFT) to the estimated features $F_q$ and $F_k$ and estimate the correlation of $F_q$ and $F_k$ in the frequency domain by:
\begin{equation}
A = \mathcal{F}^{-1}\left(\mathcal{F}(F_q)\overline{\mathcal{F}(F_k)}\right),
\label{eq: fft-attnetion}
\end{equation}
where $\mathcal{F}(\cdot)$ denotes the FFT, $\mathcal{F}^{-1}(\cdot)$ denotes the inverse FFT, and $\overline{\mathcal{F}(\cdot)}$ denotes the conjugate transpose operation.
Finally, we estimate the aggregated feature by:
\begin{equation}
V_{att} = \mathcal{L}\left(A\right)F_v,
\label{eq: fft-transformer}
\end{equation}
where a layer norm $\mathcal{L}(\cdot)$ is used to normalize $A$.
Finally, we generate the output feature of FSAS by:
\begin{equation}
X_{att} = X + \mathrm{Conv}_{1\times 1}(V_{att}),
\label{eq: fft-transformer}
\end{equation}
where $\mathrm{Conv}_{1\times 1}(\cdot)$ denotes a convolution with filter size of $1\times 1$ pixel.
The detailed network architecture of the proposed FSAS is shown in Figure~\ref{fig: Network}(b).

\subsection{Discriminative frequency domain-based FFN}
%
The FFN is used to improve the features by the scaled dot-product attention. Thus, it is important to develop an effective FFN to generate the features that facilitate the latent clear image reconstruction.
As not all the low-frequency information and high-frequency information help latent clear image restoration, we develop a DFFN that can adaptively determine which frequency information should be preserved.
However, how to effectively determine which frequency information is important. Motivated by the JPEG compression algorithm, we introduce a learnable quantization matrix $\mathbf{W}$ and learn it by an inverse method of JPEG compression to determine which frequency information should be preserved.
The proposed DFFN can be formulated by:
\begin{equation}
\begin{split}
&X_1 = \mathrm{Conv}_{1\times 1}(\mathcal{L}(X_{att}))\\
&X_1^f = \mathcal{F}(\mathcal{P}(X_1))\\
&X_2 = \mathcal{F}^{-1}(\mathbf{W}X_1^f)\\
&X_{out} = \mathcal{G}\left(\mathcal{P}^{-1}(X_2)\right)+X_{att},
\label{eq: attention_v}
\end{split}
\end{equation}
where $\mathcal{P}(\cdot)$ and $\mathcal{P}^{-1}(\cdot)$ denote the patch unfolding and folding operations in the JPEG compression method; $\mathcal{G}$ denotes GEGLU function by~\cite{GLU}.
%
The detailed network architecture of the proposed DFFN is shown in Figure~\ref{fig: Network}(c).

\subsection{Asymmetric encoder-decoder network}
\label{section:Asymmetric encoder-decoder network}
We embed the proposed FSAS and DFFN into a network based on an encoder and decoder architecture.
We note that most existing methods usually use symmetric architectures in the encoder and decoder modules.
For example, if the FSAS and DFFN are used in the encoder module, they are also used in the decoder module.
We note that the features extracted by encoder module are shallow ones, which usually contain blur effects compared to the deep features from the decoder module.
However, the blur usually changes similarity of two similar patches from clear features. Thus, using the FSAS in the encoder module may not estimate the similarity correctly, which accordingly affects image restoration.
To overcome this problem, we embed the FSAS into the decoder module, which leads to an asymmetric architecture for better image deblurring.
%
%
Figure~\ref{fig: Network}(a) shows the network architecture of the proposed asymmetric encoder-decoder network.

Finally, given a blurred image $B$, the restored image $I$ is estimated by the asymmetric encoder-decoder network:
\begin{equation}
\begin{split}
I = \mathcal{N}(B) + B,
\label{eq: latent-image}
\end{split}
\end{equation}
where $\mathcal{N}$ denotes the asymmetric encoder-decoder network.

\begin{figure*}[t]
\footnotesize
\centering
    \begin{tabular}{cccc}
    \includegraphics[width=0.24\textwidth]{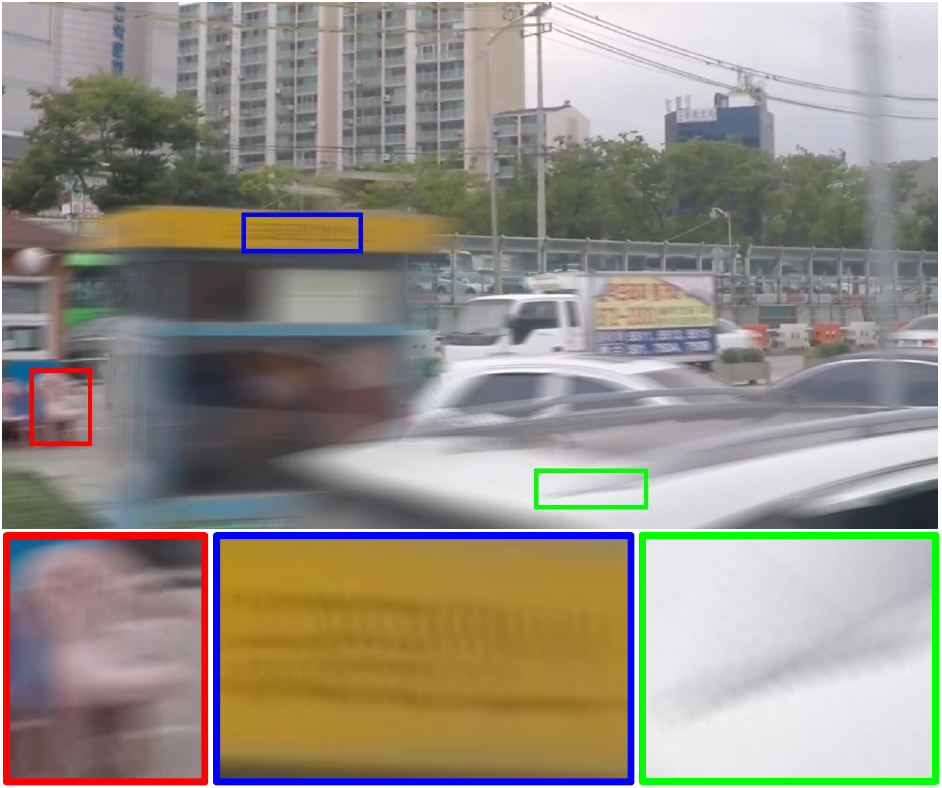}&\hspace{-4.5mm}
    \includegraphics[width=0.24\textwidth]{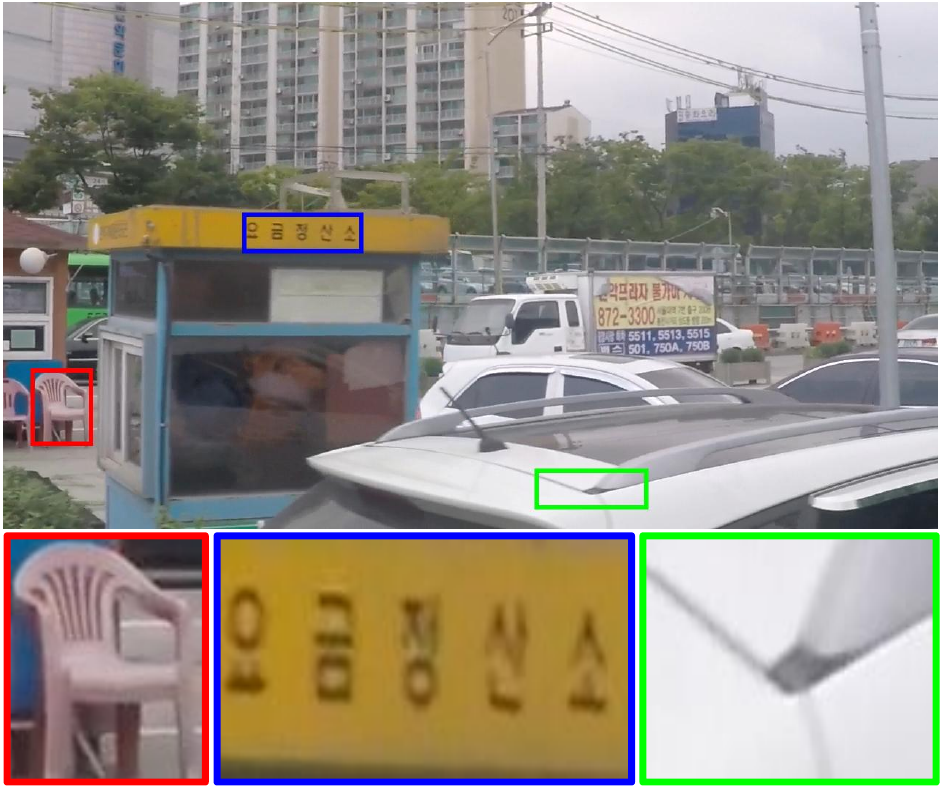}&\hspace{-4.5mm}
    \includegraphics[width=0.24\textwidth]{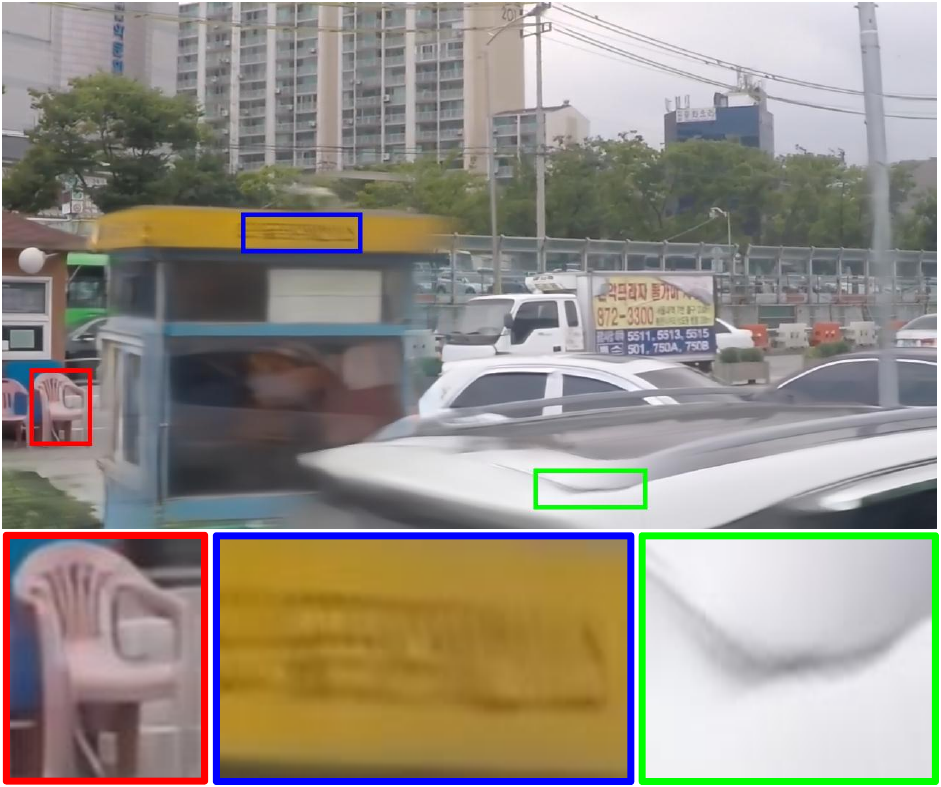}&\hspace{-4.5mm}
    \includegraphics[width=0.24\textwidth]{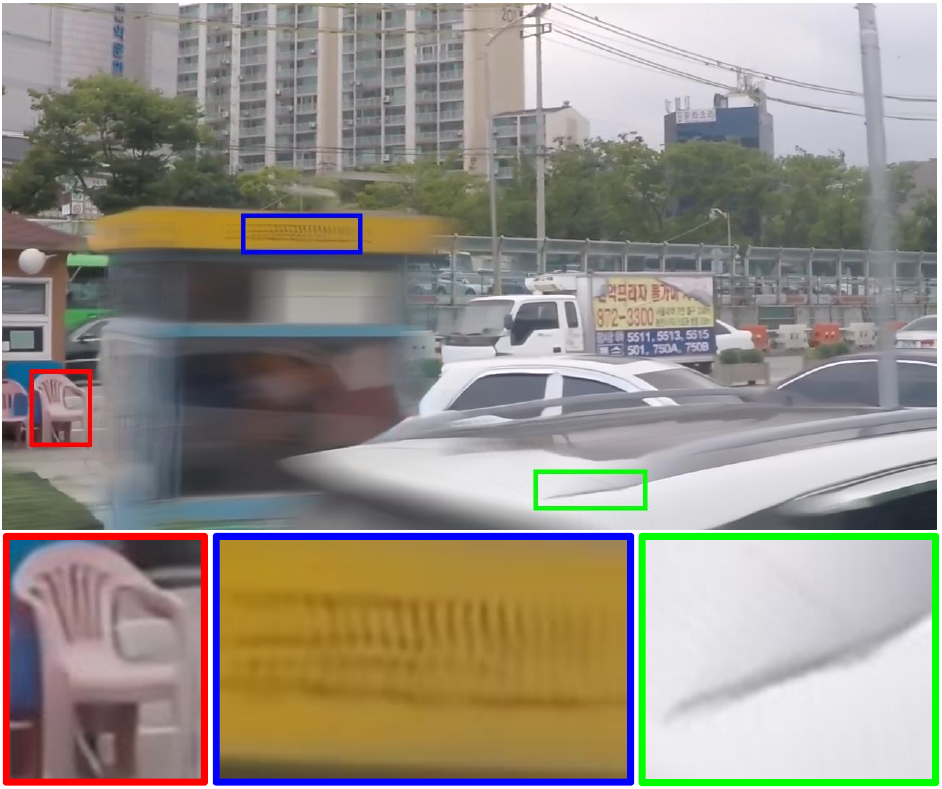}\\
    \hspace{-4.5mm}(a) Blurred image &\hspace{-4.5mm} (b) GT &\hspace{-4.5mm}(c) MIMO-Unet+~\cite{MIMO} &\hspace{-4.5mm}(d) Restormer~\cite{Restormer}\\
    \includegraphics[width=0.24\textwidth]{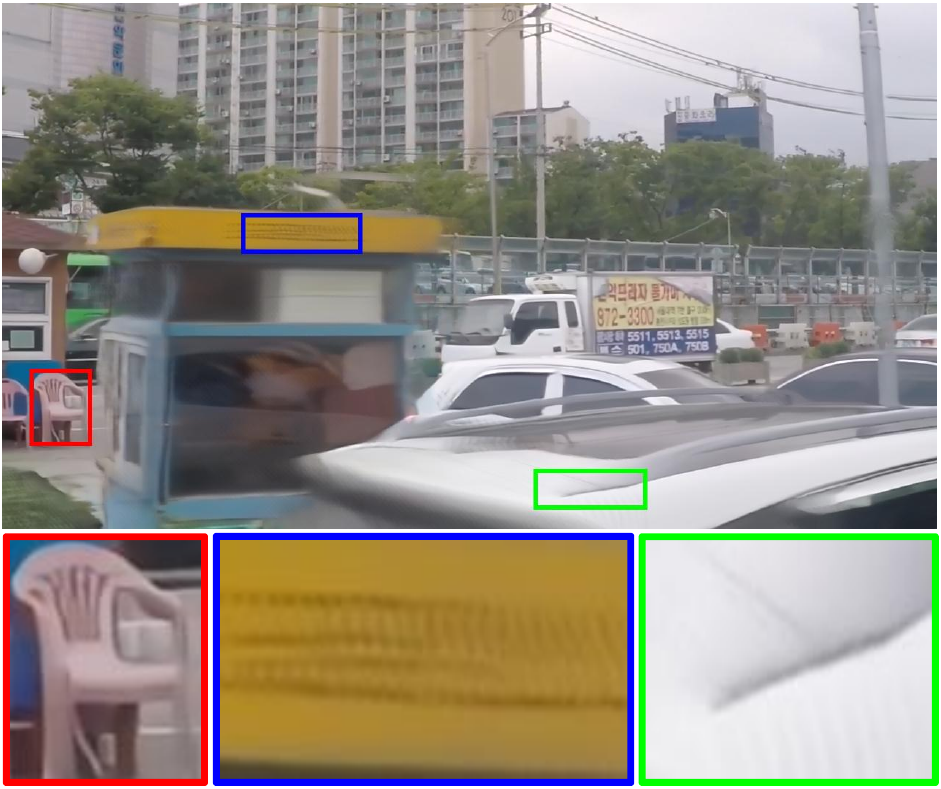} &\hspace{-4.5mm}
    \includegraphics[width=0.24\textwidth]{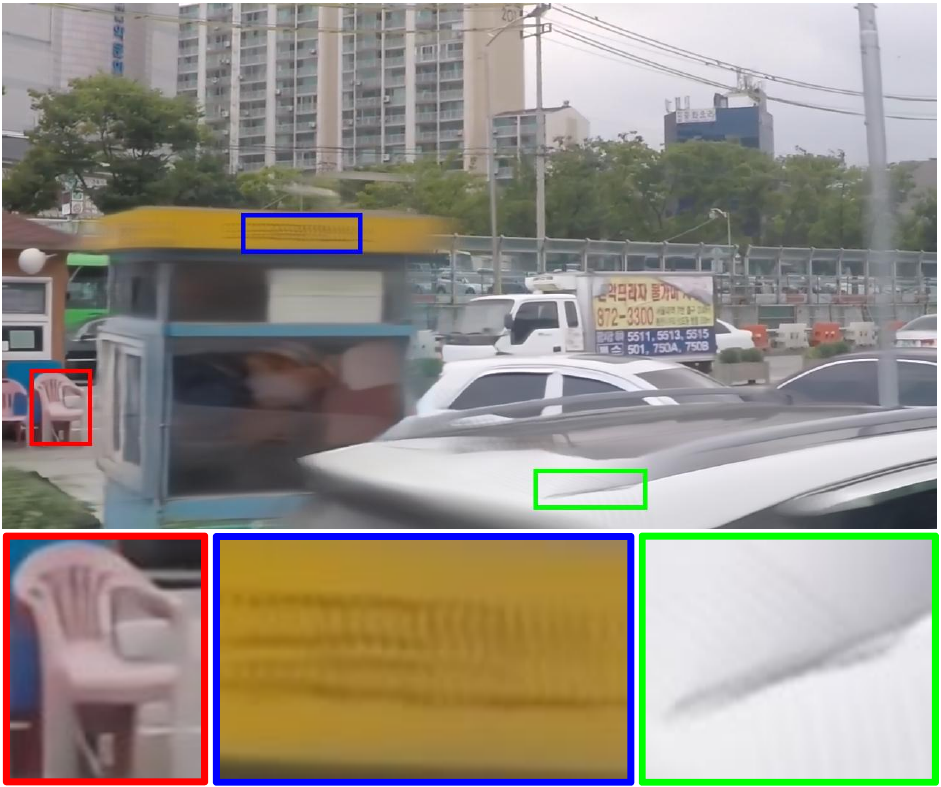} &\hspace{-4.5mm}
    \includegraphics[width=0.24\textwidth]{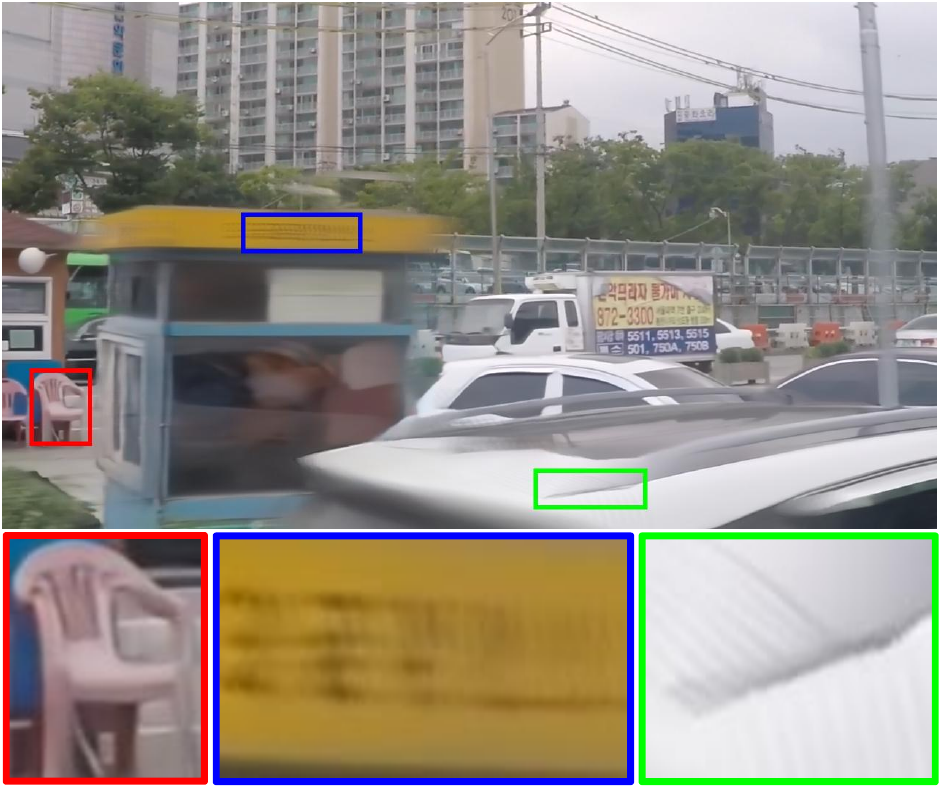} &\hspace{-4.5mm}
    \includegraphics[width=0.24\textwidth]{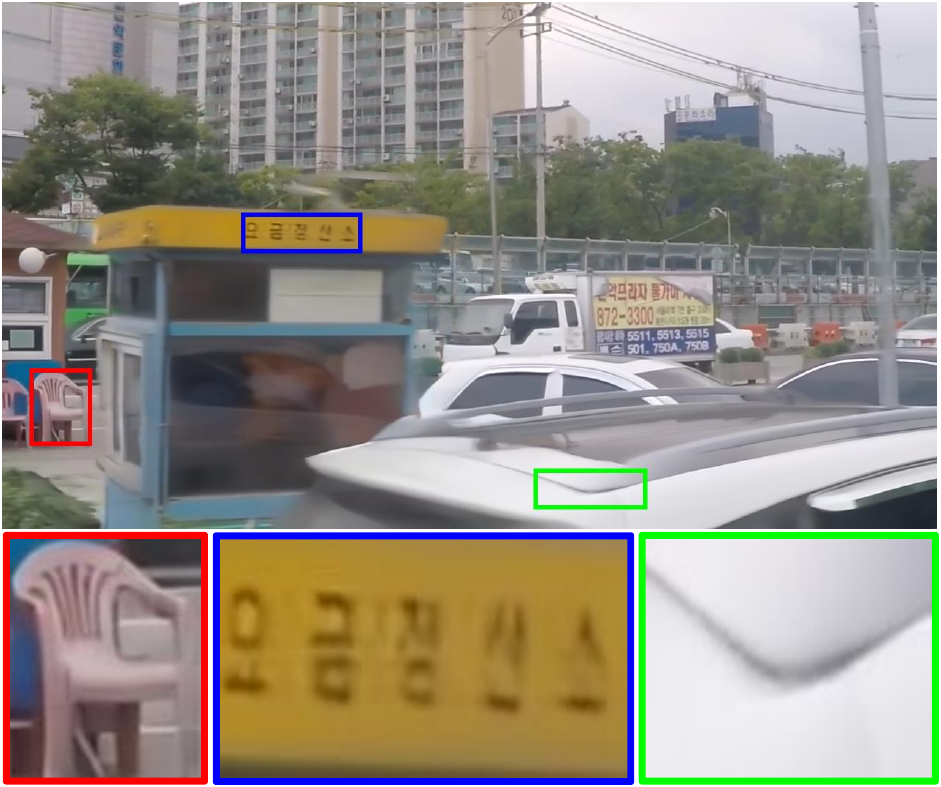}\\
\hspace{-4.5mm} (e) Stripformer~\cite{Stripformer}  &\hspace{-4.5mm} (f) Restormer-local~\cite{TLC} &\hspace{-4.5mm} (g) NAFNet~\cite{NAFNet} &\hspace{-4.5mm} (h) Ours
    \end{tabular}
\vspace{-2mm}
    \caption{Deblurred results on the GoPro dataset~\cite{GoPro}. The deblurred results in (c)-(g) still contain significant blur effects. The proposed method generates a clearer image. For example, the characters and boundaries are much clearer.}
    \label{fig:GoPro_result}
\vspace{-4mm}
\end{figure*}

\section{Experimental Results}
\label{sec:experiment}
In this section, we evaluate the proposed method and compare it with state-of-the-art methods using public benchmark datasets.
\subsection{Datasets and parameter settings}
{\flushleft \textbf{Datasets.}}
We evaluate our method on commonly used image deblurring datasets including the GoPro dataset~\cite{GoPro}, the HIDE dataset~\cite{HIDE}, and the RealBlur dataset~\cite{Realblur}.
We follow the protocols of existing methods for fair comparisons.
{\flushleft \textbf{Parameter settings.}}
We use the same loss function as~\cite{MIMO} to constrain the network and train it using the Adam~\cite{Adam} optimizer with default parameters.
The initial value of the learning rate is $10^{-3}$ and is updated with the cosine annealing strategy after 600,000 iterations. The minimum value of the learning rate is $10^{-7}$.
The patch size is empirically set to be $256\times 256$ pixels and the batch size is set to be 16.
We adopt the same data augmentation method as~\cite{Restormer} during the training.
The patch size for the weight matrix estimation is empirically set to be $8\times 8$ based on the JPEG compression method.
Similarity, we also use the patch of the size $8\times 8$ pixels when computing the self-attention~\eqref{eq: fft-attnetion}.

Due to the page limit, we include more experimental results in the supplemental material. The training code and models will be available to the public.
\begin{table}[!t] \footnotesize
\centering
    \caption{Quantitative evaluations on the GoPro dataset\cite{GoPro}.}
    \label{tab:GoPro}
    \vspace{-2mm}
    \begin{tabular}{*{2}{l}*{10}{c}}
        \toprule
       Methods & PSNRs &  SSIMs & Parameters (M) & \\
        \midrule
        DeblurGAN-v2~\cite{DeblurGANv2}      & 29.55 & 0.9340  & 60.9 \\
        SRN~\cite{SRN}                       & 30.26 & 0.9342 & 6.8  \\
        DMPHN~\cite{DMPHN}                   & 31.20 & 0.9453 & 21.7 \\
        SAPHN~\cite{SAPHN}                   & 31.85 & 0.9480 & 23.0 \\
        MIMO-Unet+~\cite{MIMO}               & 32.45 & 0.9567 & 16.1 \\
        MPRNet~\cite{MPRNet}                 & 32.66 & 0.9589 & 20.1 \\
        DeepRFT+~\cite{Deeprft}              & 33.23 & 0.9632 & 23.0\\
        Restormer~\cite{Restormer}           & 32.92 & 0.9611 & 26.1 \\
        Uformer-B~\cite{Uformer}             & 33.06 & 0.9670 & 50.9 \\
        Stripformer~\cite{Stripformer}       & 33.08 & 0.9624 & 19.7 \\
        MPRNet-local~\cite{TLC}              & 33.31 & 0.9637 & 20.1 \\
        Restormer-local~\cite{TLC}           & 33.57 & 0.9656 & 26.1 \\
        NAFNet~\cite{NAFNet}                 & 33.71 & 0.9668 & 67.9 \\
        Ours                                 & \textbf{34.21} & \textbf{0.9692} & 16.6 \\
        \bottomrule
    \end{tabular}
  \vspace{-5mm}
\end{table}

\subsection{Comparisons with the state of the arts}
We compare our method with state-of-the-art ones and use the PSNR and SSIM to evaluate the quality of restored images.
\vspace{-2mm}
{\flushleft \textbf{Evaluations on the GoPro dataset.}}
We first evaluate our method on the commonly used GoPro dataset by~\cite{GoPro}.
For fair comparisons, we follow the protocols of this dataset and retrain or fine-tune the deep learning methods that are not trained on this dataset.
Table~\ref{tab:GoPro} shows the quantitative evaluation results. Our method generates the results with the highest PSNR and SSIM values.
Compared to the state-of-the-art CNN-based methods, NAFNet~\cite{NAFNet}, the PSNR gain of our method is at least 0.5dB higher than NAFNet, while the number of the proposed model parameters is a quarter of the NAFNet.
In addition, compared to the Transformer-based methods~\cite{Restormer,Uformer,Stripformer}, our method has the fewest model parameters while the performance is better.
%

Figure~\ref{fig:GoPro_result} shows visual comparisons of the proposed method and the evaluated ones on the GoPro dataset.
As demonstrated by~\cite{Restormer}, the CNN-based methods~\cite{MIMO,NAFNet} do not effectively explore non-local information for latent clear image restoration. Therefore, the deblurred results by the methods~\cite{MIMO,NAFNet} still contain significant blur effect as shown in Figure~\ref{fig:GoPro_result}(c) and (g).
The Transformer-based methods~\cite{Restormer,Stripformer,TLC} are able to model the global contexts for image deblurring. However, some main structures, e.g., characters and chairs, are not recovered well (see Figure~\ref{fig:GoPro_result}(d)-(f)).

In contrast to existing Transformer-based methods that are based on the spatial domain, we develop an efficient frequency domain-based Transformer, where the proposed DFFN is able to discriminately estimate useful frequency information for latent clear image restoration. Thus, the deblurred results contain clear structures, and the characters are much clearer as shown in Figure~\ref{fig:GoPro_result}(h).

\begin{table}[!t]\footnotesize
\centering
    \caption{Quantitative evaluations on the RealBlur
    dataset~\cite{Realblur} in terms of PSNR and SSIM.}
    \label{tab:Realblur}
    \vspace{-2mm}
    \begin{tabular}{lcccc}
        \toprule
                & \multicolumn{2}{c}{Realblur-R} &   \multicolumn{2}{c}{Realblur-J}                    \\
        Methods                         & ~PSNRs~  &  ~SSIMs~ & ~PSNRs~ & ~SSIMs~ \\
        \midrule
        DeblurGAN-v2~\cite{DeblurGANv2} & 36.44 & 0.9347 & 29.69 & 0.8703 \\
        SRN~\cite{SRN}                  & 38.65 & 0.9652 & 31.38 & 0.9091 \\
        MIMO-Unet+~\cite{MIMO}          &   -   &   -    & 31.92 & 0.9190 \\
        BANet~\cite{BANet}              & 39.55 & 0.9710 & 32.00 & 0.9230 \\
        DeepRFT+~\cite{Deeprft}         & 39.84 & 0.9721 & 32.19 & 0.9305 \\
        Stripformer~\cite{Stripformer}  & 39.84 & 0.9737 & 32.48 & 0.9290 \\
        Ours                    &\textbf{40.11} &\textbf{0.9753} &\textbf{32.62} &\textbf{0.9326 }\\
        \bottomrule
    \end{tabular}
   \vspace{-5mm}
\end{table}

\begin{figure*}[t]
\footnotesize
\centering
    \begin{tabular}{cccc}
    \includegraphics[width=0.24\textwidth, height = 0.29\linewidth]{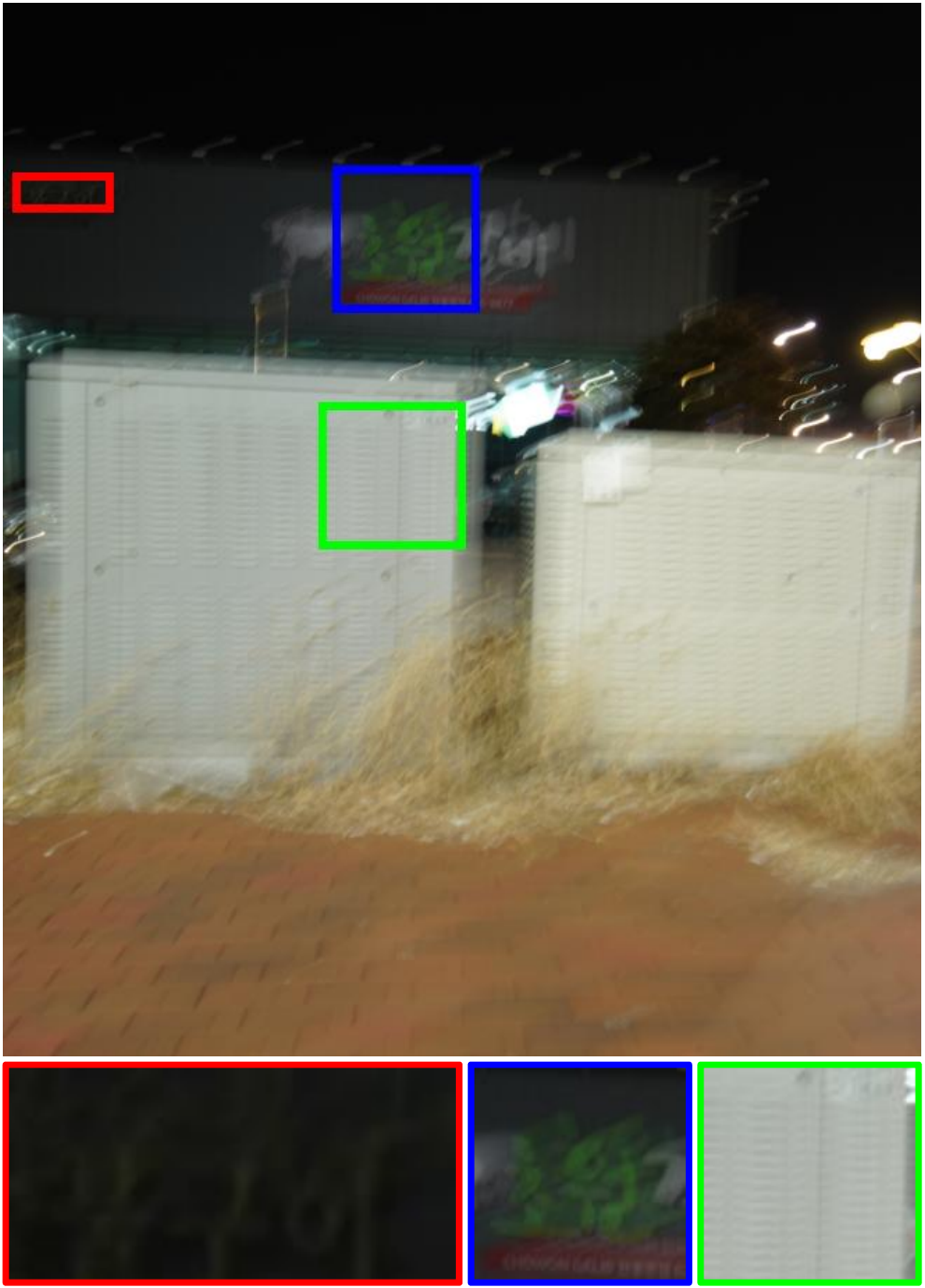} &\hspace{-4.5mm}
    \includegraphics[width=0.24\textwidth, height = 0.29\linewidth]{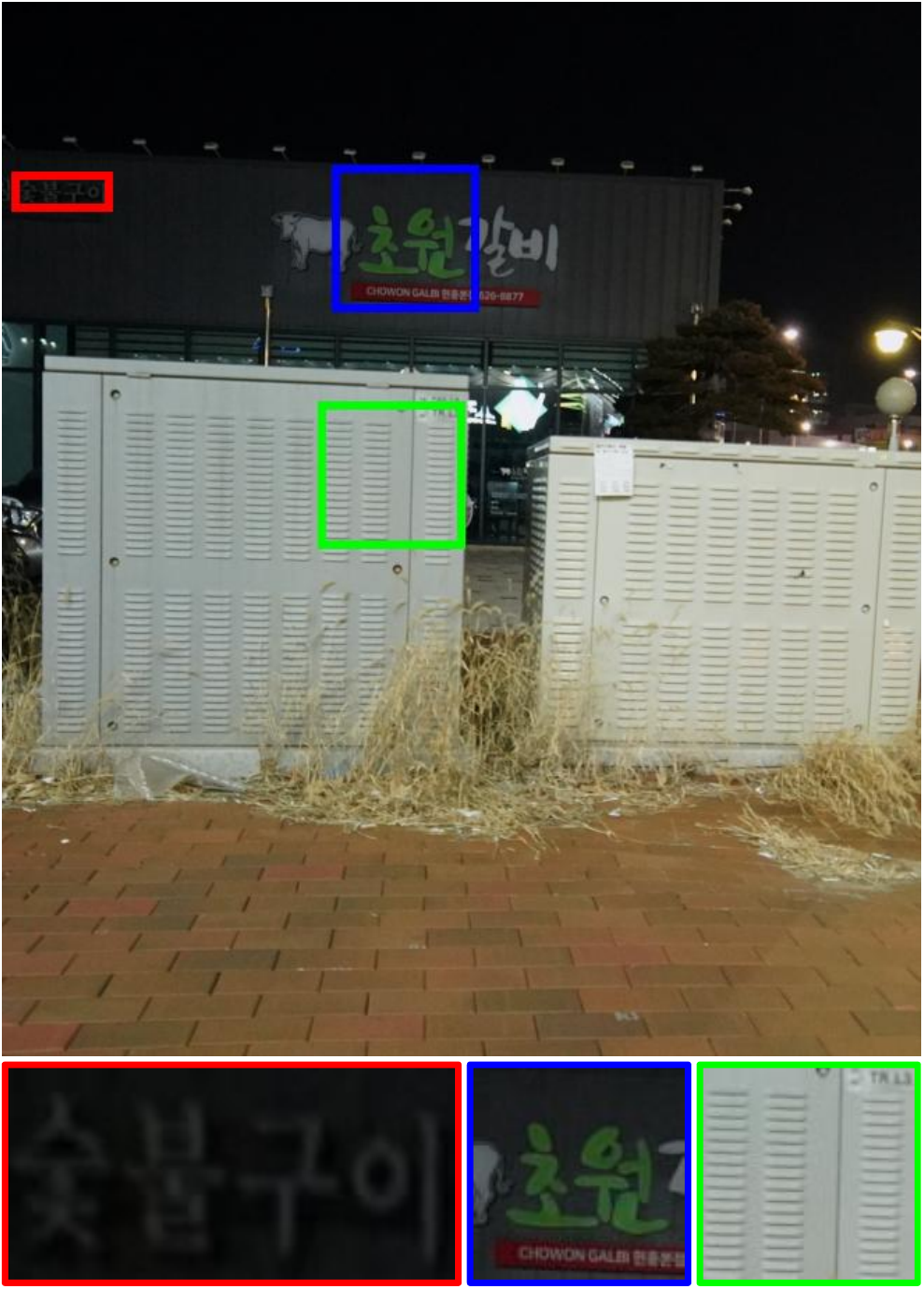} &\hspace{-4.5mm}
    \includegraphics[width=0.24\textwidth, height = 0.29\linewidth]{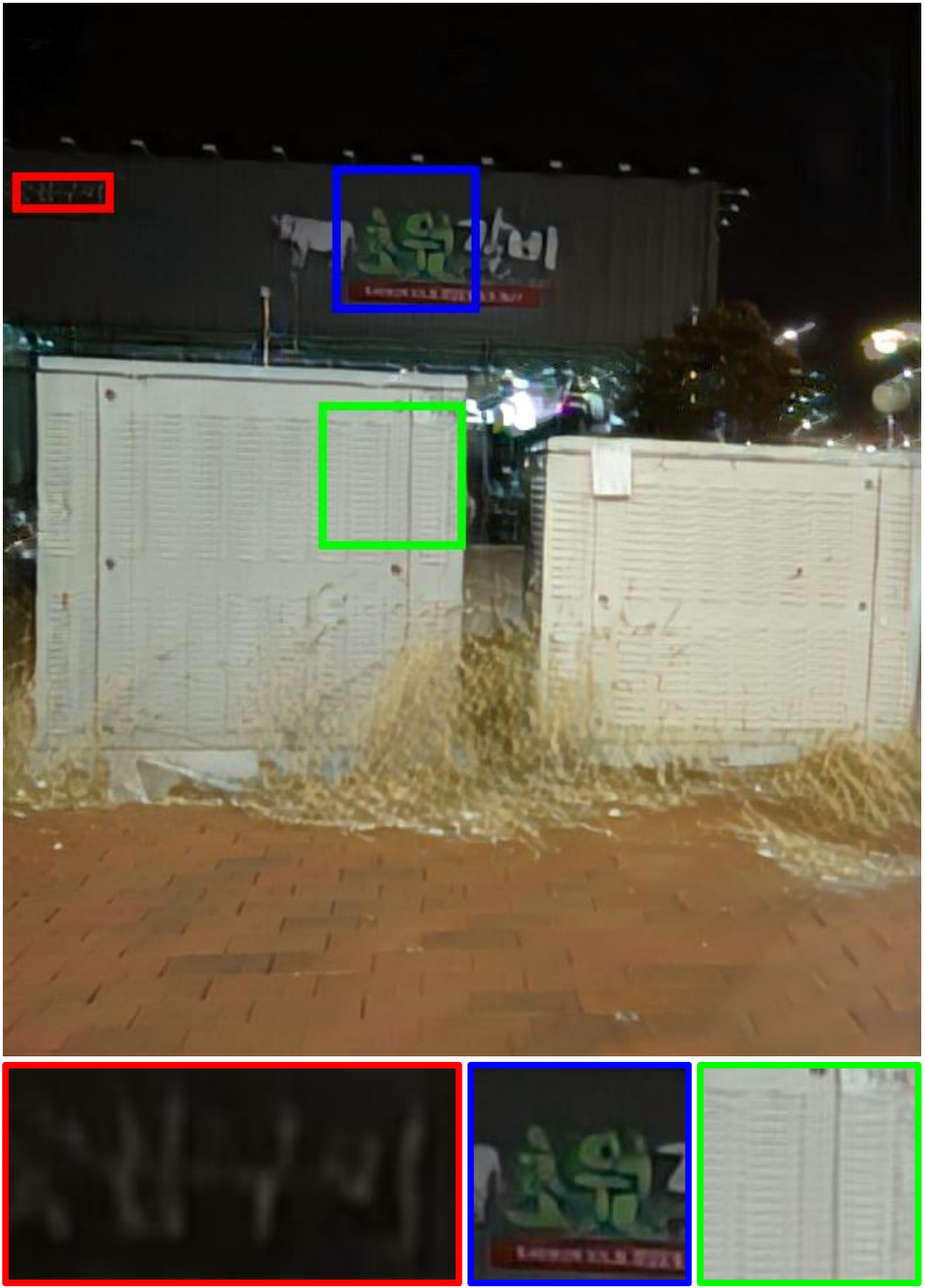} &\hspace{-4.5mm}
    \includegraphics[width=0.24\textwidth, height = 0.29\linewidth]{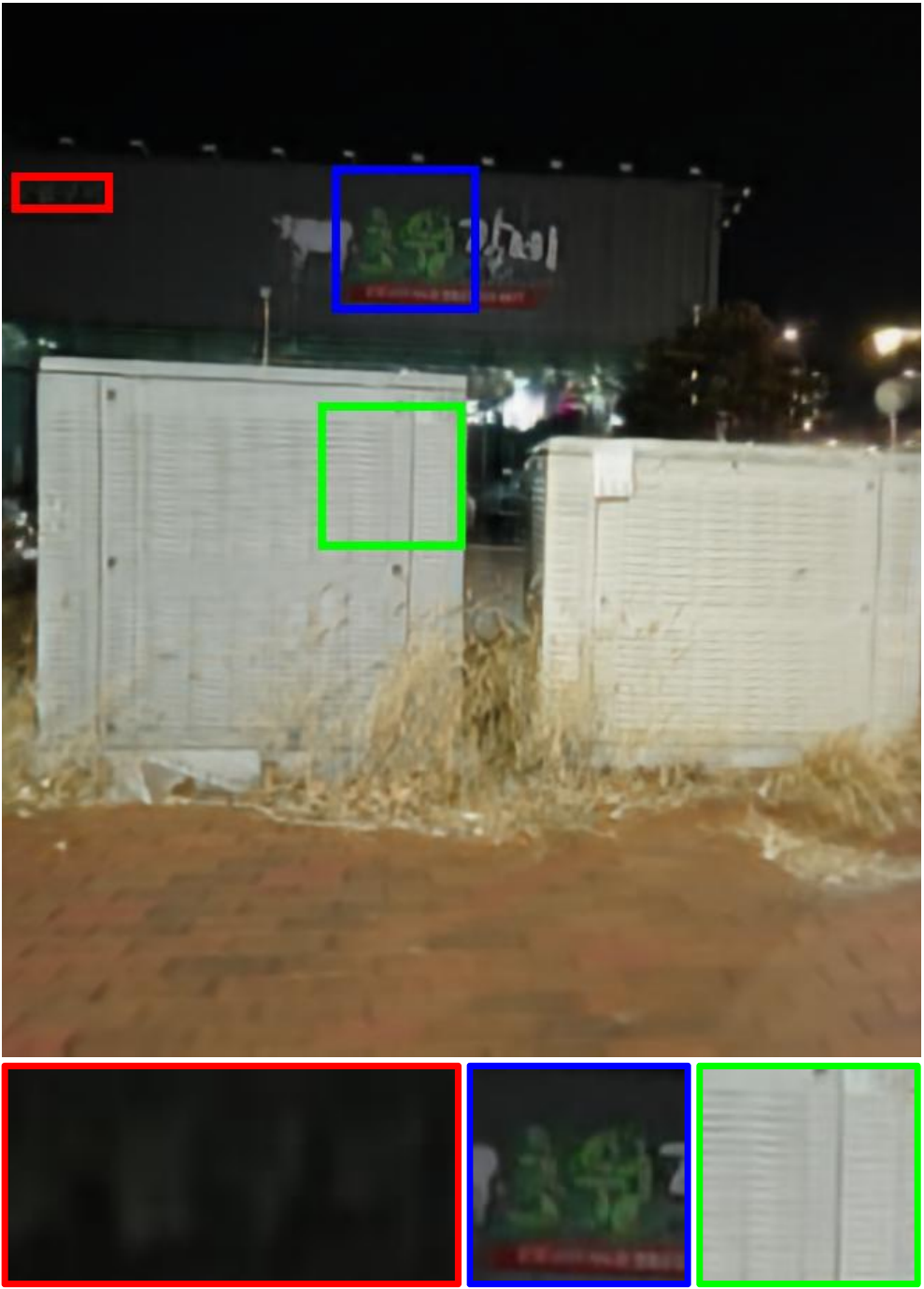} \\
\hspace{-4.5mm} (a) Blurred image &\hspace{-4.5mm} (b) GT &\hspace{-4.5mm} (c) DeblurGAN~\cite{DeblurGAN} &\hspace{-4.5mm} (d) SRN~\cite{SRN}\\
    \includegraphics[width=0.24\textwidth, height = 0.29\linewidth]{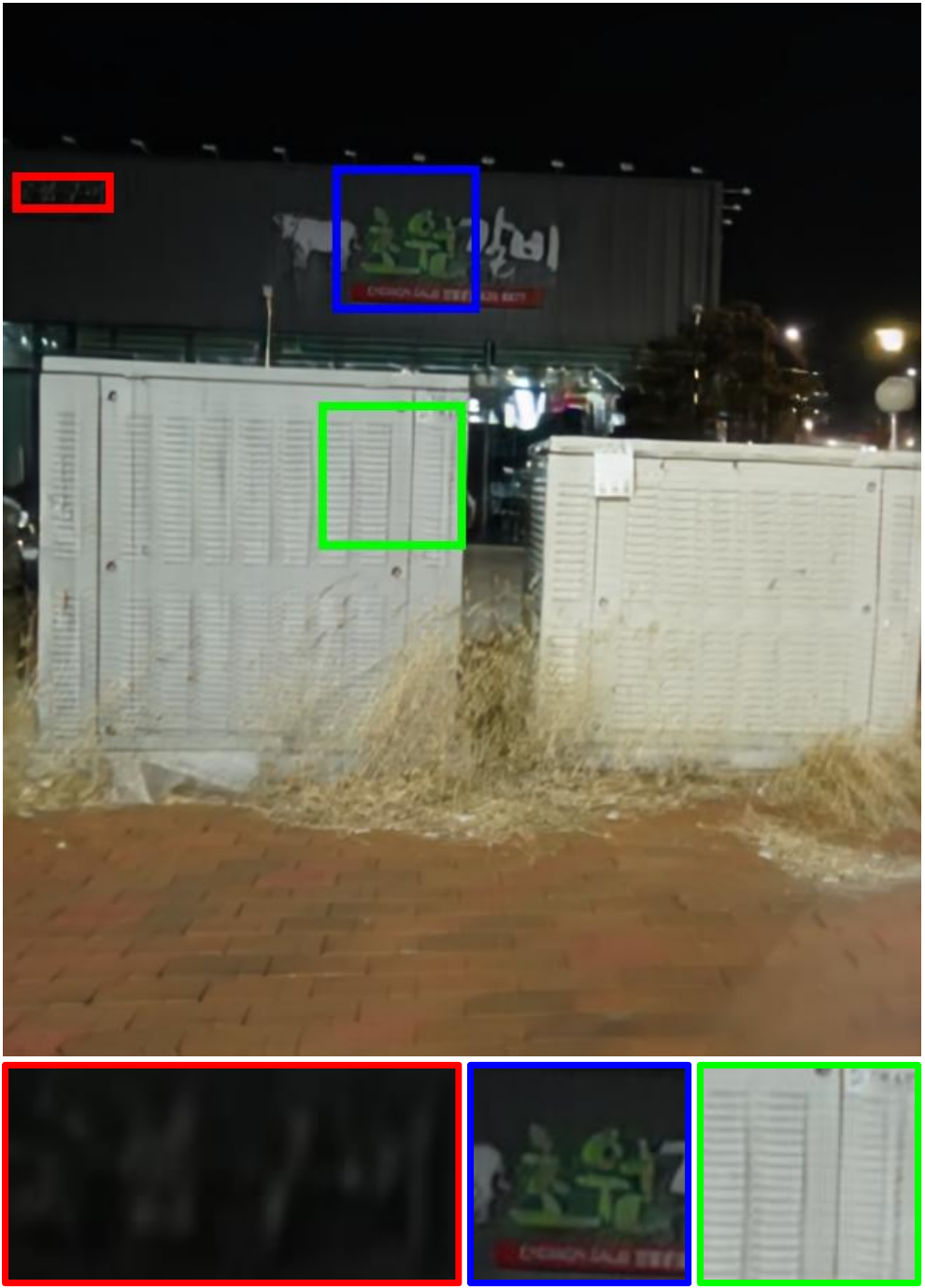}  &\hspace{-4.5mm}
    \includegraphics[width=0.24\textwidth, height = 0.29\linewidth]{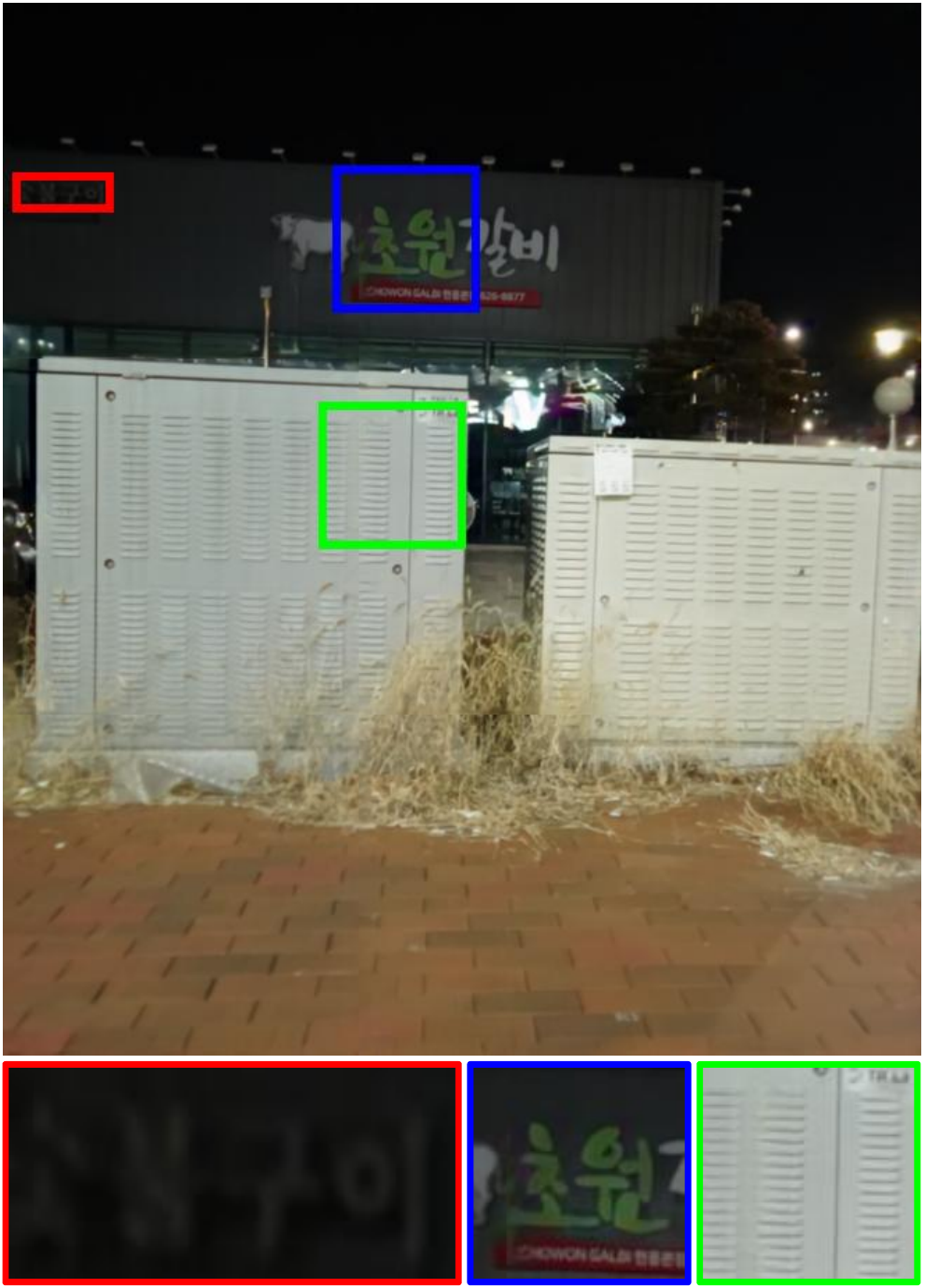}  &\hspace{-4.5mm}
    \includegraphics[width=0.24\textwidth, height = 0.29\linewidth]{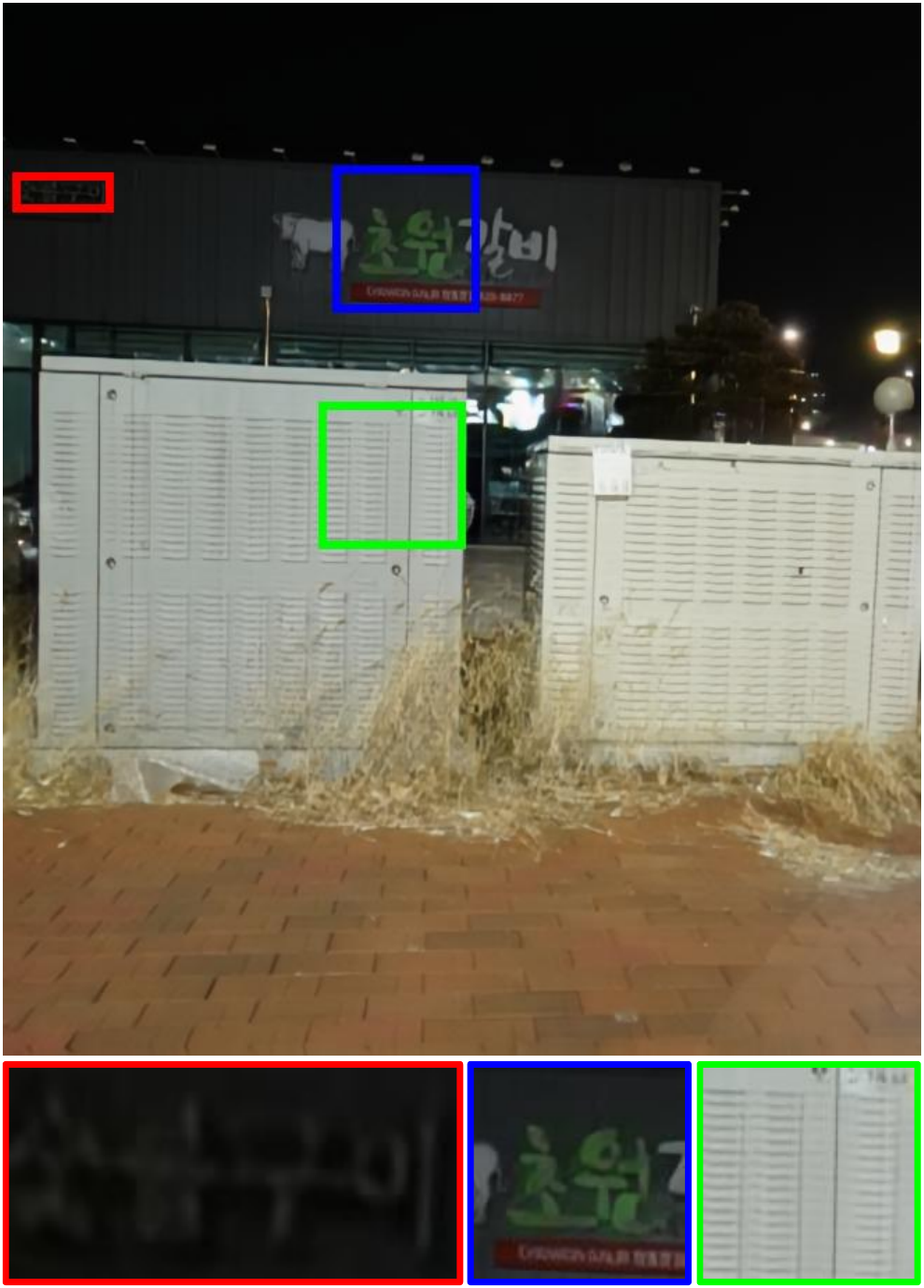}  &\hspace{-4.5mm}
    \includegraphics[width=0.24\textwidth, height = 0.29\linewidth]{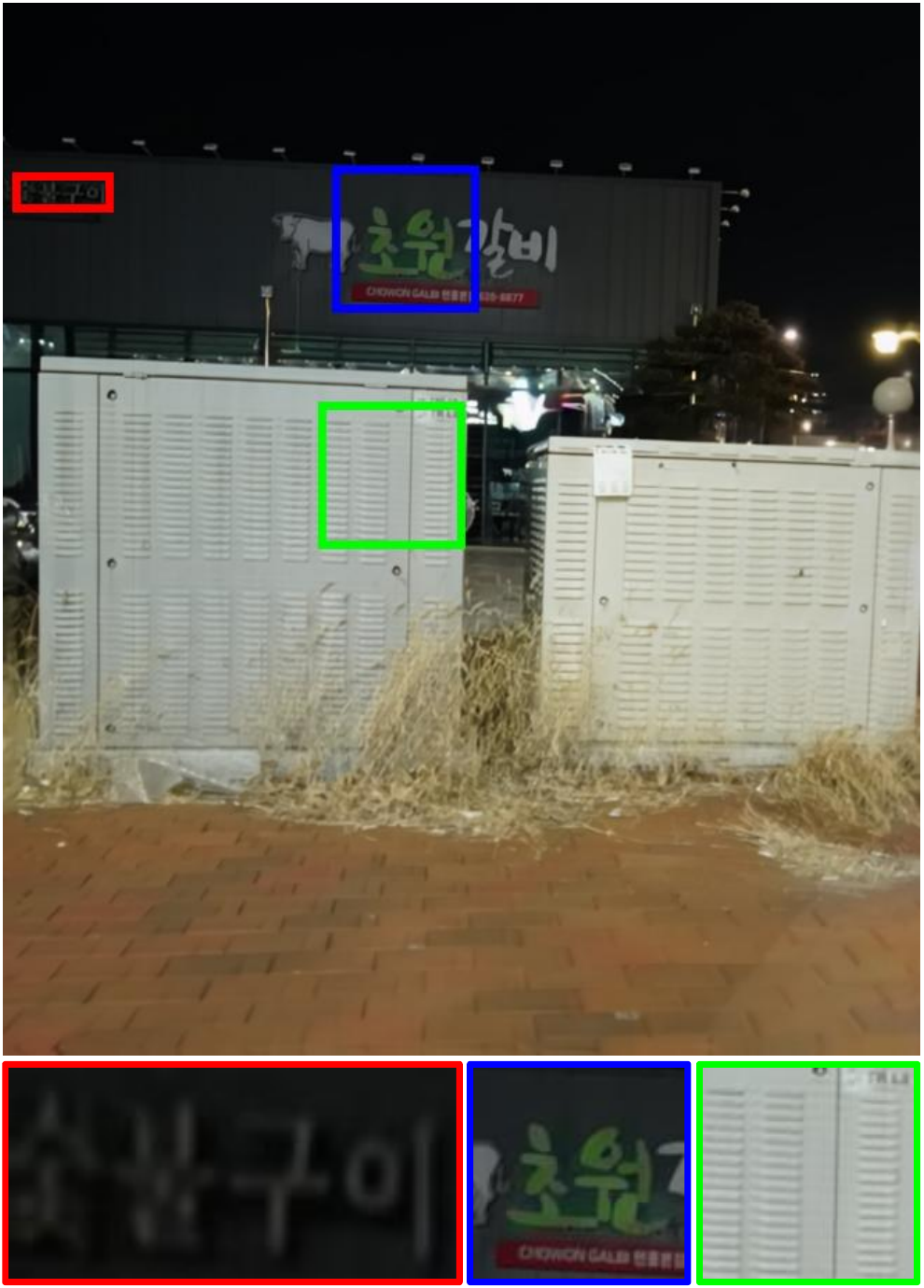}\\
\hspace{-4.5mm} (e) MIMO-Unet+~\cite{MIMO} &\hspace{-4.5mm} (f) DeepRFT+~\cite{Deeprft} &\hspace{-4.5mm} (g) Stripformer~\cite{Stripformer} &\hspace{-4.5mm} (h) Ours
    \end{tabular}
\vspace{-2mm}
    \caption{Deblurred results on the RealBlur dataset~\cite{Realblur}. The characters or the structural details in (c)-(g) are not recovered well. The proposed method generates an image with much clearer characters and structural details.}
    \label{fig:Realblur_result}
\vspace{-4mm}
\end{figure*}

\vspace{-2mm}
{\flushleft \textbf{Evaluations on the RealBlur dataset.}}
We further evaluate our method on the RealBlur dataset by~\cite{Realblur} and follow the protocols of this dataset for fair comparisons.
The test dataset of~\cite{Realblur} includes a RealBlur-R test set from the raw images and RealBlur-J test set from the JPEG images.
Table~\ref{tab:Realblur} summarizes the quantitative evaluation results on the above mentioned test sets.
The proposed method generates the results with higher PSNR and SSIM values.
%
%

Figure~\ref{fig:Realblur_result} shows the visual comparisons on the RealBlur dataset, where our method generates the results with clearer characters and finer structural details (Figure~\ref{fig:Realblur_result}(h)).

\begin{table}[!t]\footnotesize
\centering
    \caption{Quantitative evaluations on the HIDE dataset~\cite{HIDE}. We use the models trained on the GoPro dataset~\cite{GoPro}
    for fair comparisons.}
    \vspace{-2mm}
    \label{tab:HIDE}
    \begin{tabular}{lcccc}
        \toprule
       Methods                         & PSNRs & SSIMs & Parameters (M) & \\
        \midrule
        DeblurGAN-v2\cite{DeblurGANv2} & 26.61 & 0.8750 &  60.9 \\
        SRN~\cite{SRN}                  & 28.36 & 0.9040 &  6.8  \\
        DMPHN~\cite{DMPHN}              & 29.09 & 0.9240 &  21.7 \\
        SAPHN~\cite{SAPHN}              & 29.98 & 0.9300 &  23.0 \\
        MIMO-Unet+~\cite{MIMO}          & 29.99 & 0.9304 &  16.1 \\
        MPRNet~\cite{MPRNet}            & 30.96 & 0.9397 &  20.1 \\
        Stripformer~\cite{Stripformer} & 31.03 & 0.9395 &  19.7 \\
        MPRNet-local~\cite{TLC}        & 31.19 & 0.9418 &  20.1 \\
        Restormer~\cite{Restormer}      & 31.22 & 0.9423 &  26.1 \\
        NAFNet\cite{NAFNet}           & 31.31  & 0.9427 &  67.9 \\
        Restormer-local~\cite{TLC}     & 31.49 & 0.9447 &  26.1 \\
        Ours                           & \textbf{31.62} & \textbf{0.9455} & 16.6\\
        \bottomrule
    \end{tabular}
    \vspace{-5mm}
\end{table}

\begin{figure*}[t]
\footnotesize
\centering
    \begin{tabular}{cccc}
    \includegraphics[width=0.24\textwidth]{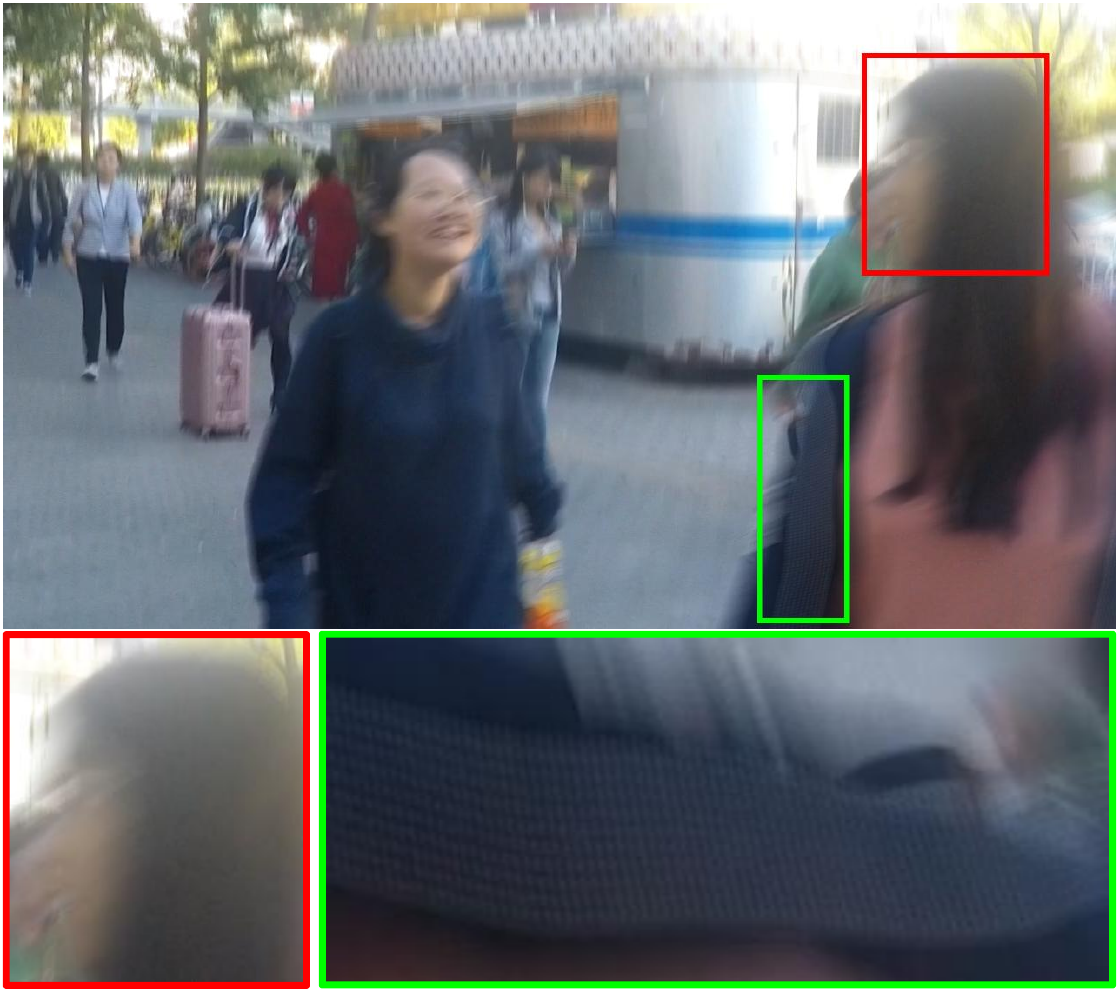}&\hspace{-4.5mm}
    \includegraphics[width=0.24\textwidth]{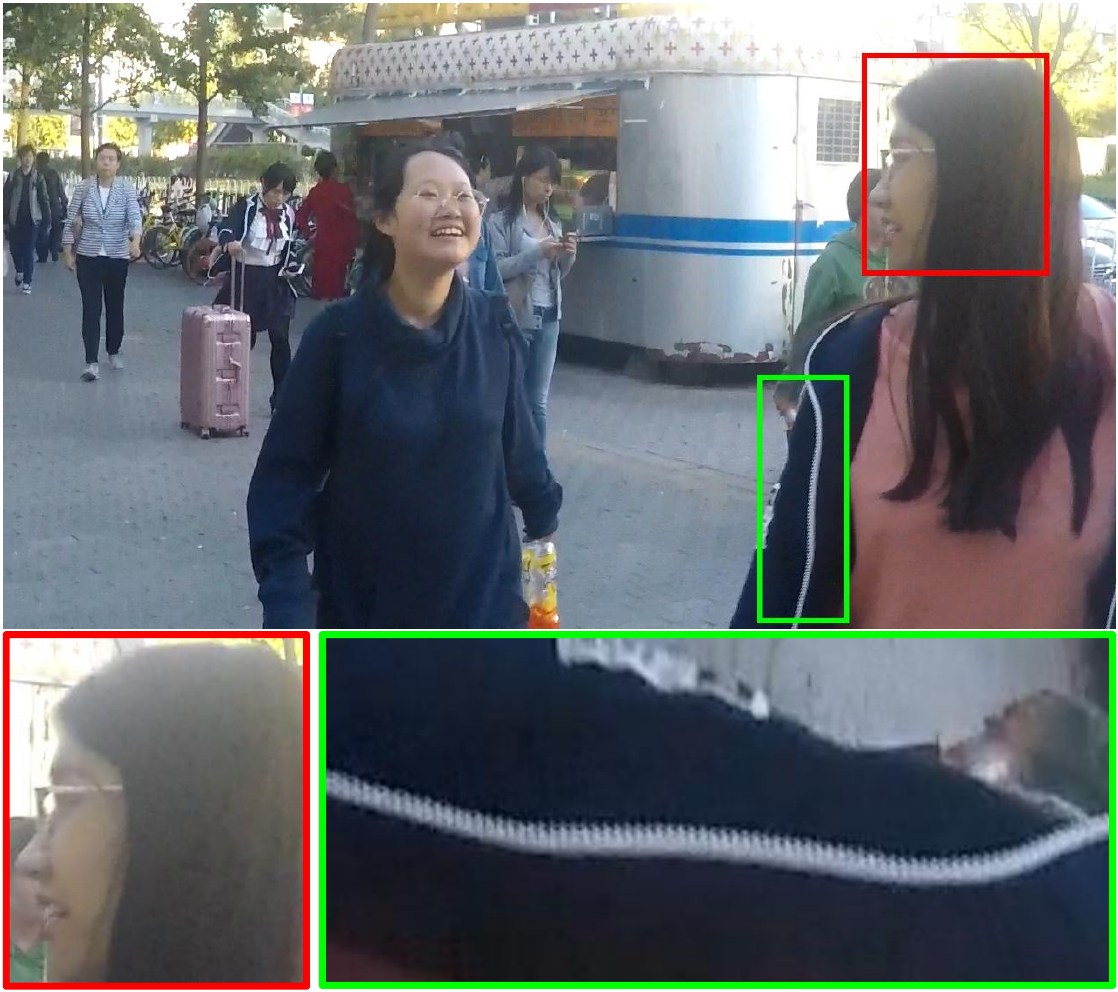}&\hspace{-4.5mm}
    \includegraphics[width=0.24\textwidth]{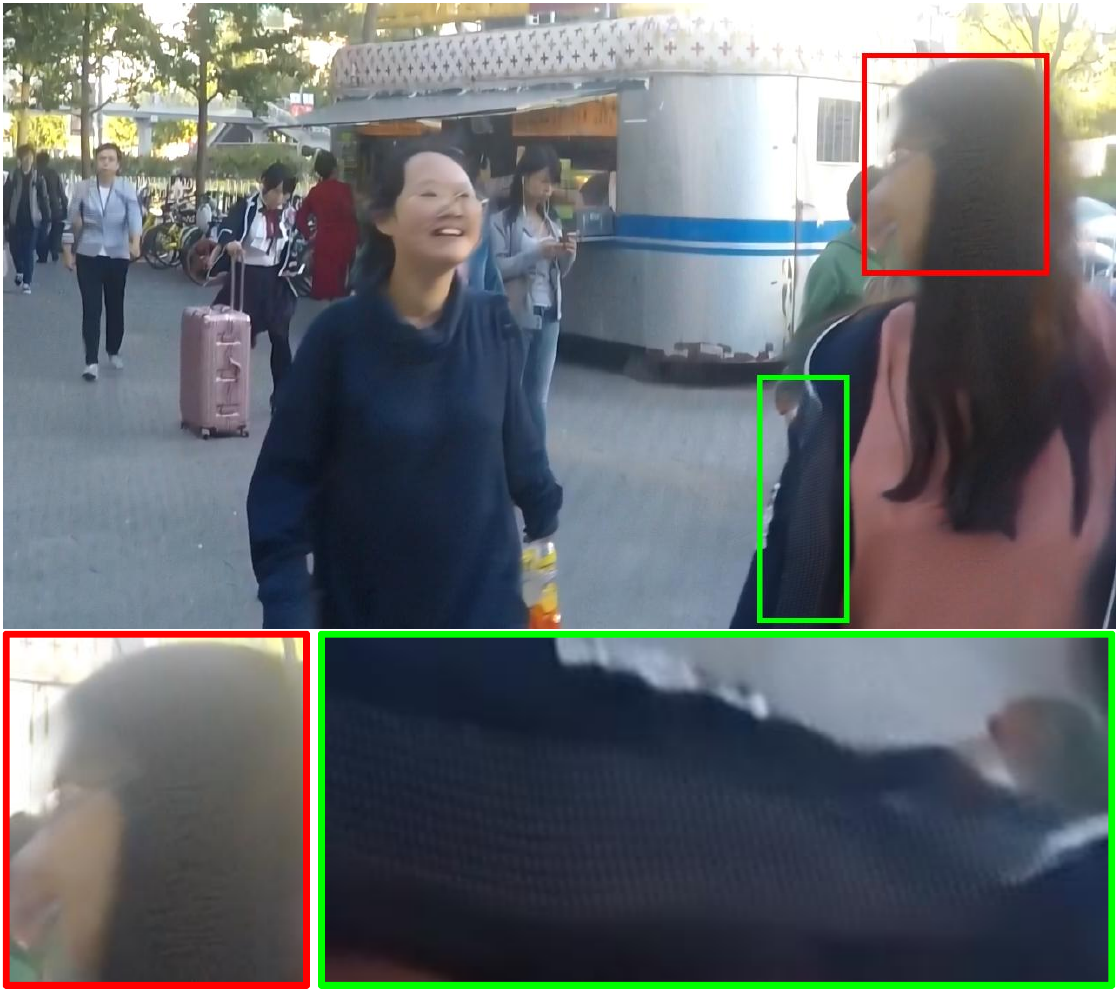}&\hspace{-4.5mm}
    \includegraphics[width=0.24\textwidth]{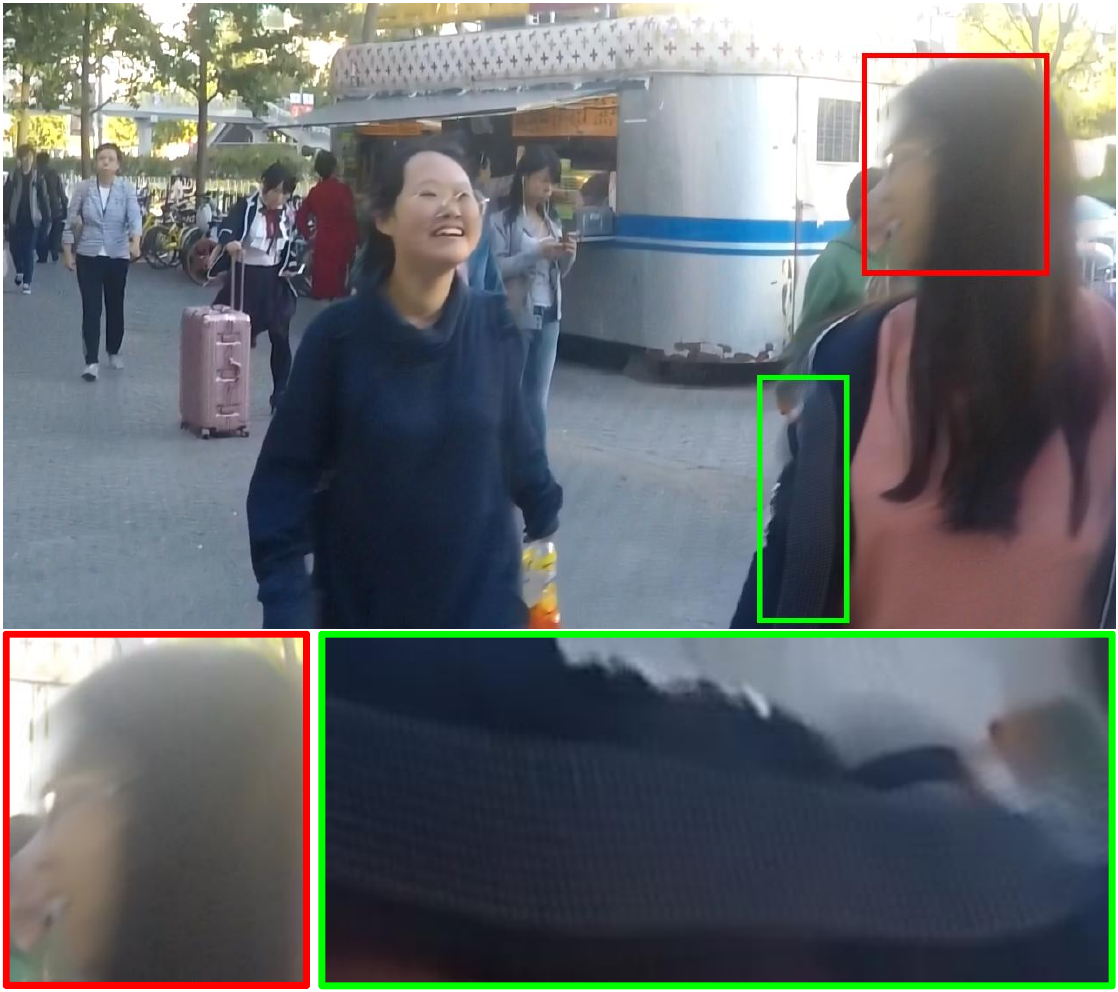}\\
    \hspace{-4.5mm}(a) Blurred image &\hspace{-4.5mm} (b) GT &\hspace{-4.5mm}(c) MPRNet~\cite{MPRNet} &\hspace{-4.5mm}(d) Restormer~\cite{Restormer}\\
    \includegraphics[width=0.24\textwidth]{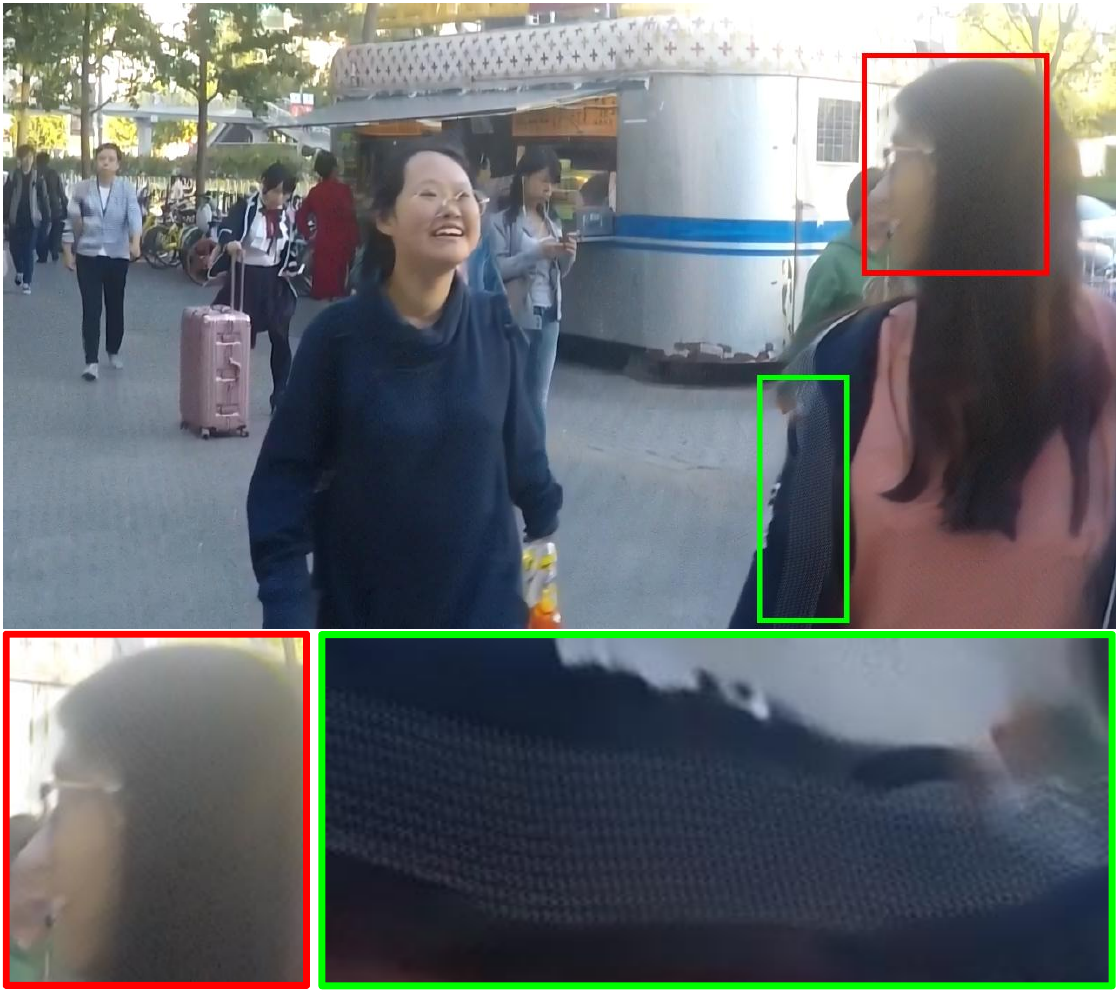} &\hspace{-4.5mm}
    \includegraphics[width=0.24\textwidth]{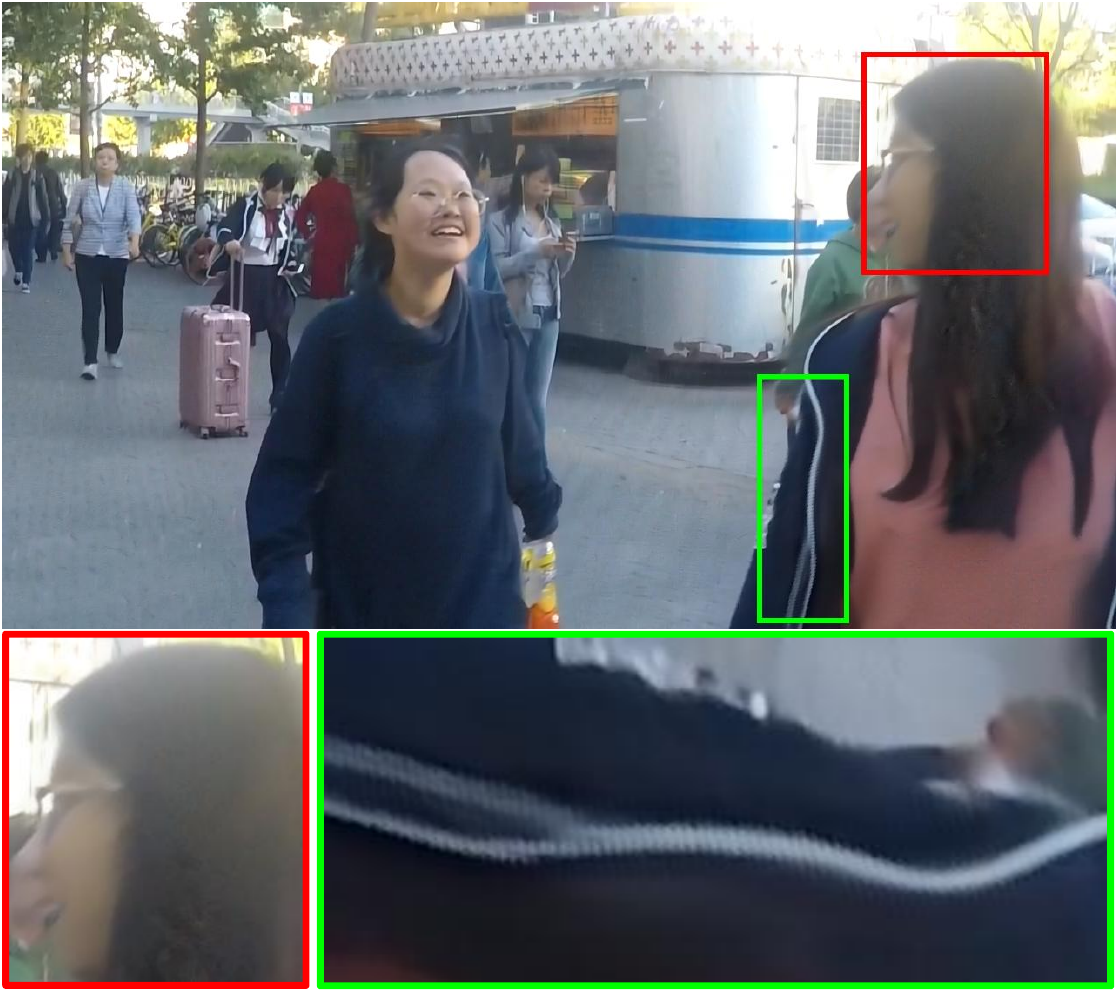} &\hspace{-4.5mm}
    \includegraphics[width=0.24\textwidth]{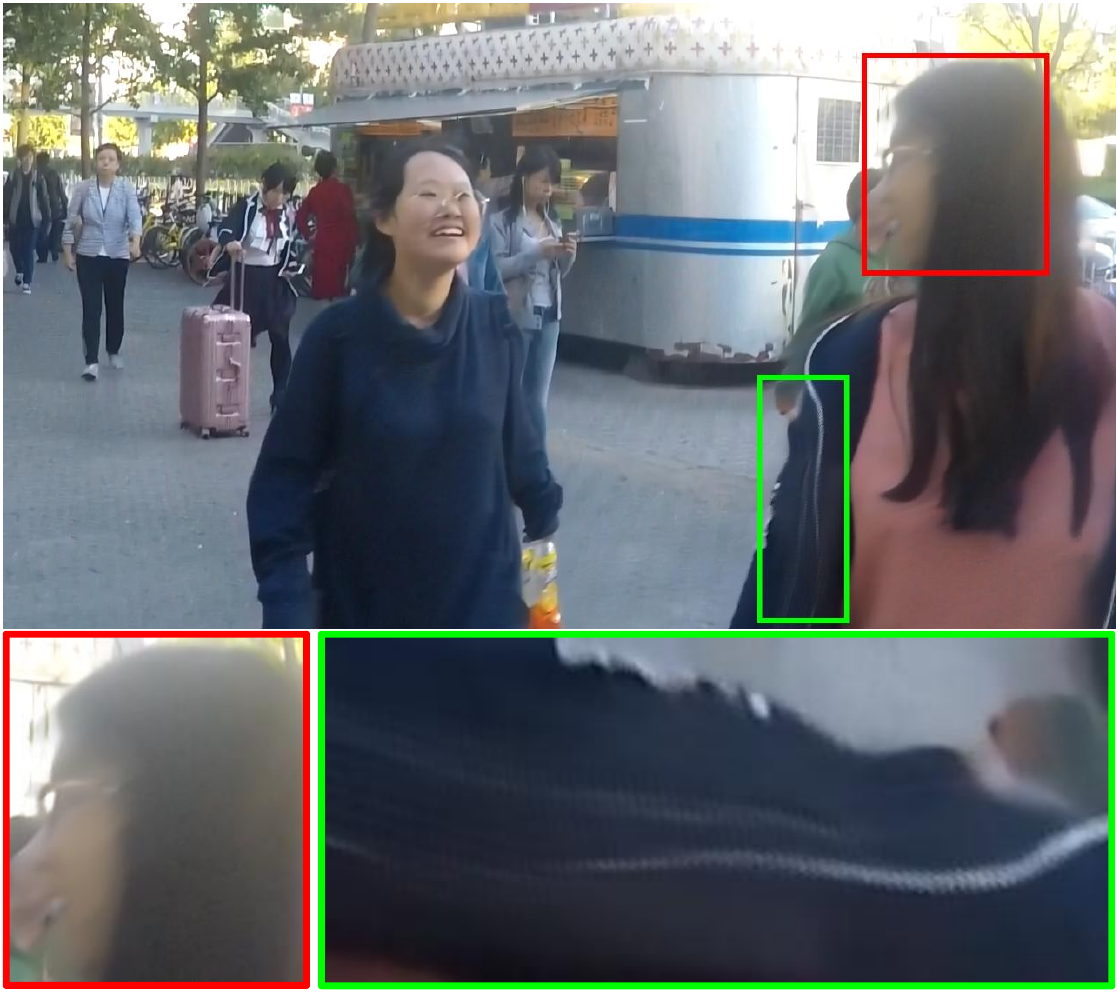} &\hspace{-4.5mm}
    \includegraphics[width=0.24\textwidth]{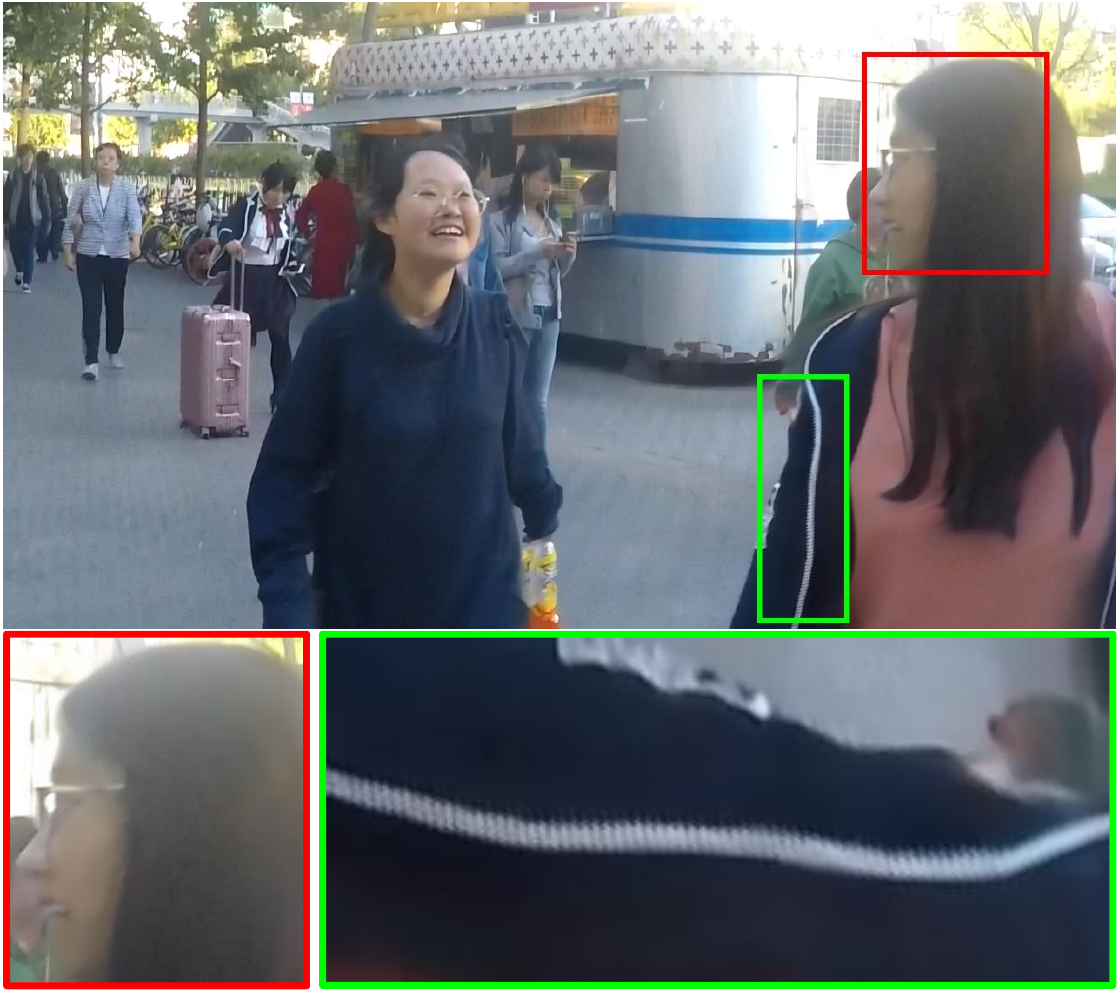}\\
\hspace{-4.5mm} (e) Stripformer~\cite{Stripformer}  &\hspace{-4.5mm} (f) NAFNet~\cite{NAFNet} &\hspace{-4.5mm} (g) Restormer-local~\cite{TLC} &\hspace{-4.5mm} (h) Ours
    \end{tabular}
\vspace{-2mm}
    \caption{Deblurred results on the HIDE dataset~\cite{HIDE}. The deblurred results in (c)-(g) still contain significant blur effects. The proposed method generates much clearer images.}
    \label{fig:HIDE_result}
\vspace{-4mm}
\end{figure*}
\vspace{-2mm}
{\flushleft \textbf{Evaluations on the HIDE dataset.}}
We then evaluate our method on the HIDE dataset~\cite{HIDE}, which mainly contains humans.
Similar to state-of-the-art methods~\cite{MPRNet,MIMO}, we directly use the models of the evaluated methods, which are trained on the GoPro dataset for test.
Table~\ref{tab:HIDE} shows that the quality of the deblurred images generated by the proposed method is better than the evaluated methods, suggesting that our method has a better generalization ability as models are not trained on this dataset.

We show some visual comparisons in Figure~\ref{fig:HIDE_result}. We note that the evaluated methods do not recover the humans well. In contrast, our method generates better images. For example, the faces and zipper of clothes are much clearer.

\begin{table}[!t]
  \caption{Memory and running time comparisons of the Transformer based on our method and the window methods~\cite{Swin,Uformer}. The size of the test image is $1280\times 720$ pixels. The test environment is based on a machine with an NVIDIA GeForce RTX 3090 GPU.
  ``\#GPU Mem.'' denotes the maximum GPU memory consumption that is computed by the ``torch.cuda.max memory allocated()'' function. ``\#Avg. Time'' denotes the average running time.
  }
   \vspace{-3mm}
   \label{tab: space-time-complexity}
\footnotesize
\resizebox{0.49\textwidth}{!}{
 \centering
 \begin{tabular}{lcccc}
        \toprule
                & \multicolumn{2}{c}{Window-based method~\cite{Uformer}} &   \multicolumn{2}{c}{Ours}                    \\
        Window size                         &\#Avg. Time &\#GPU Mem. & \#Avg. Time&\#GPU Mem. \\
        \midrule
        $8\times 8$                             & 53ms & 6.3G & 44ms & 6.5G \\
        $16\times 16$                           & 56ms & 7.1G & 44ms & 6.2G \\
        $32\times 32$                           & 89ms & 12.0G & 43ms & 6.0G \\
        $64\times 64$                           & - & Out of memory  & 42ms & 5.9G \\
        $1280\times 720$                        & - &Out of memory& 42ms & 5.9G \\
        \bottomrule
    \end{tabular}
}
\vspace{-3mm}
\end{table}

\begin{table}[!t]
  \caption{Quantitative evaluations of each component in the proposed method on the GoPro dataset~\cite{GoPro}.
  }
   \vspace{-3mm}
   \label{tab: effect-of-fsas}
\footnotesize
\resizebox{0.48\textwidth}{!}{
 \centering
 \begin{tabular}{lcccc|cc}
    \toprule
                          &     FSAS     &   Swin attention   &    FFN     &     DFFN    &    PSNRs/SSIMs\\
    \hline
  w/ only FFN             &\XSolidBrush   &\XSolidBrush     &\CheckmarkBold      &\XSolidBrush      &33.19/0.9626\\
  w/ only DFFN            &\XSolidBrush   &\XSolidBrush     &\XSolidBrush      &\CheckmarkBold    &33.55/0.9651 \\
  SA w/ SD                &\XSolidBrush   &\CheckmarkBold   &\XSolidBrush      &\CheckmarkBold    &33.46/0.9645 \\
  FSAS+FFN                &\CheckmarkBold &\XSolidBrush     &\CheckmarkBold    &\XSolidBrush      &33.61/0.9654\\
  FSAS+DFFN               &\CheckmarkBold  &\XSolidBrush     &\XSolidBrush     &\CheckmarkBold      &33.73/0.9663\\
 \bottomrule
  \end{tabular}
}
\vspace{-5mm}
\end{table}

\vspace{-1mm}
\section{Analysis and Discussion}
\vspace{-1mm}
\label{sec:Ablation}
We have shown that exploring the properties of Transformers in the frequency domain generates favorable results against state-of-the-art methods.
In this section, we provide deeper analysis on the proposed method and demonstrate the effect of the main components.
For the ablation studies in this section, we train our method and all the baselines on the GoPro dataset using the batch size of $8$ to illustrate the effect of each component in our method.
\vspace{-2mm}
{\flushleft \textbf{Effect of FSAS.}}
The proposed FSAS is used to reduce the computational cost. According to the properties of FFT, the space and time complexity of the FSAS are $O(N)$ and $O(NC\log N)$,
which are much lower than $O(N^2)$ and $O(N^2C)$ in the original computation of the scaled dot-product attention, where $C$ is the number of features.
We further examine the space and time complexity of the FSAS and the window-based strategy~\cite{Swin,Uformer} for Transformers.
Table~\ref{tab: space-time-complexity} shows that using the proposed FSAS needs a small GPU memory and is much more efficient compared to the window-based strategy~\cite{Uformer}.

Moreover, as the proposed FSAS is performed in the frequency domain, one may wonder whether the scaled dot-product attention estimated in the spatial domain performs better or not.
To answer this question, we compare the FSAS with the baseline method that performs in the spatial domain (SA w/ SD for short).
As the space complexity of the original scaled dot-product attention is $O(N^2)$,
it is not affordable to train ``SA w/ SD'' when using the same settings as the proposed FSAS. We use the Swin Transformer~\cite{Swin} for comparison as it is much more efficient.
Table~\ref{tab: effect-of-fsas} shows the quantitative evaluation results on the GoPro dataset.
The method that computes the scaled dot-product attention in the spatial domain does not generate good deblurred results, where its PSNR value is 0.27 lower (see comparisons of ``SA w/ SD'' and ``FSAS+DFFN'' in Table~\ref{tab: effect-of-fsas}).
The main reason is that although using the shifted window partitioning method reduces the computational cost, it does not fully explore the useful information across different windows.
In contrast, the space complexity of the proposed FSAS is $O(N)$ and does not need the shifted window partitioning as an approximation, thus leading to better deblurred results.
\begin{figure}[!t]\footnotesize
\centering
\begin{tabular}{ccc}
\hspace{-3mm}
\includegraphics[width=0.156\textwidth]{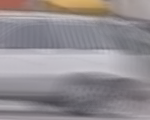} &\hspace{-4mm}
\includegraphics[width=0.156\textwidth]{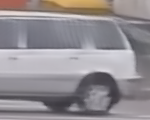} &\hspace{-4mm}
\includegraphics[width=0.156\textwidth]{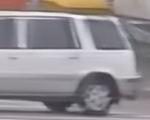} \\
\hspace{-3mm}(a) Blurred image  &\hspace{-4mm}  (b) Spatial domain  &\hspace{-4mm} (c) Frequency domain\\
\end{tabular}
\vspace{-3mm}
\caption{Effectiveness of the scaled dot-product attention computation in the spatial and frequency domains. Computing the scaled dot-product attention using the FSAS in the frequency domain generates a clearer image.}
\label{fig: FSAS-domain}
\vspace{-1mm}
\end{figure}
Figure~\ref{fig: FSAS-domain}(b) further shows that using the shifted window partitioning method as an approximation of scaled dot-product attention in the spatial domain does not remove blur effectively. In contrast, the proposed FSAS generates clearer images.

\begin{figure}[!t]\footnotesize
\centering
\begin{tabular}{ccc}
\hspace{-3mm}
\includegraphics[width=0.156\textwidth]{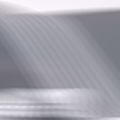} &\hspace{-4mm}
\includegraphics[width=0.156\textwidth]{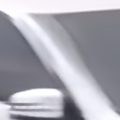} &\hspace{-4mm}
\includegraphics[width=0.156\textwidth]{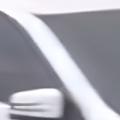} \\
\hspace{-3mm}(a) Blurred image  &\hspace{-4mm}  (b) Ours w/o FSAS  &\hspace{-4mm} (c) Ours\\
\end{tabular}
\vspace{-3mm}
\caption{Effectiveness of the proposed FSAS on image deblurring. Using the proposed FSAS generates a clearer image.}
\label{fig: FSAS-with-without}
\vspace{-2mm}
\end{figure}

Moreover, compared to the baseline method only using FFN (``w/ only FFN''), using the proposed FSAS in this baseline generates much better results, where the PSNR value is 0.42dB higher (see comparisons of ``w/ only FFN'' and ``FSAS+FFN'' in Table~\ref{tab: effect-of-fsas}). The visual comparisons in Figure~\ref{fig: FSAS-with-without}(b) and (c) further demonstrate that using the proposed FSAS facilitates the blur removal well, where the boundaries are recovered well as shown in Figure~\ref{fig: FSAS-with-without}(c).

\begin{figure}[!t]\footnotesize
\centering
\begin{tabular}{cccc}
\hspace{-3mm}
\includegraphics[width=0.156\textwidth]{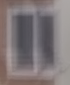} &\hspace{-4mm}
\includegraphics[width=0.156\textwidth]{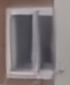} &\hspace{-4mm}
\includegraphics[width=0.156\textwidth]{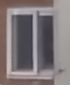} \\
\hspace{-3mm}(a) Blurred image &\hspace{-4mm} (b) w/ only FFN &\hspace{-4mm} (c) w/ only DFFN\\
\end{tabular}
\vspace{-3mm}
\caption{Effectiveness of the proposed DFFN on image deblurring. }
\label{fig: dffn}
\vspace{-5mm}
\end{figure}

\vspace{-2mm}
{\flushleft \textbf{Effect of DFFN.}}
The proposed DFFN is used to discriminatively estimate useful frequency information for latent clear image restoration.
To demonstrate its effectiveness on image deblurring, we compare the proposed method with two baselines. For the first baseline, we compare the proposed method only using the DFFN  (w/ only DFFN for short) and the proposed method only using the original FFN (w/ only FFN for short). For the second baseline, we compare the proposed method with the one that replaces the DFFN with the original FFN in the proposed method (FSAS+FFN).
The comparisons of ``w/ only DFFN'' and ``w/ only FFN'' in Table~\ref{tab: effect-of-fsas} show that using the proposed DFFN generates better results, where the PSNR value is 0.36dB higher.

In addition, the comparisons of ``FSAS+FFN'' and ``FSAS+DFFN'' in Table~\ref{tab: effect-of-fsas} show that using the proposed DFFN further improves the performance.

Figure~\ref{fig: dffn} shows the visualization results by these above mentioned baseline methods. Using the proposed DFFN generates better deblurred images, where the windows are recovered well shown in Figure~\ref{fig: dffn}(c).

\begin{figure}[!t]\footnotesize
\centering
\begin{tabular}{cccc}
\hspace{-4mm}
\includegraphics[width=0.156\textwidth]{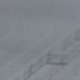} &\hspace{-4mm}
\includegraphics[width=0.156\textwidth]{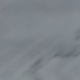} &\hspace{-4mm}
\includegraphics[width=0.156\textwidth]{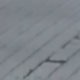}\\
\hspace{-4mm} (a) Blurred image &\hspace{-4mm} (b) FSAS in enc\&dec  &\hspace{-4mm} (c) FSAS in dec (Ours)\\
\end{tabular}
\vspace{-3mm}
\caption{Effectiveness of the asymmetric encoder-decoder network on image deblurring.}
\label{fig: netwotk-encoder}
\vspace{-4mm}
\end{figure}

\vspace{-2mm}
{\flushleft \textbf{Effect of the asymmetric encoder-decoder network.}}
As demonstrated in Section~\ref{section:Asymmetric encoder-decoder network}, the shallow features extracted by encoder module usually contain blur effects that affect the estimations of FSAS.
We thus embed it into the decoder module, which leads to an asymmetric encoder-decoder network for better image deblurring.
To examine the effect of this network design, we compare the network that puts the FSAS into both the encoder and decoder modules (``FSAS in enc\&dec'' in Table~\ref{tab: fsas-location}).
Table~\ref{tab: fsas-location} shows that using the FSAS in the decoder module generates better results, where the PSNR value is at least 0.17dB higher.
The visual comparisons in Figure~\ref{fig: netwotk-encoder}(b) and (c) further demonstrate that using the FSAS in the decoder module generates better clear images.

\begin{table}[!t]
  \caption{Quantitative evaluations of the asymmetric encoder-decoder network on the GoPro dataset.
  }
   \vspace{-3mm}
   \label{tab: fsas-location}
\footnotesize
 \centering
 \begin{tabular}{lcccc}
    \toprule
    Methods       &~~~~~~FSAS in enc\&dec~~~~~~  &~~~~~~FSAS in dec (Ours)~~~~~~ \\
    \hline
 PSNRs            &33.56          & \bf{33.73}  \\
 SSIMs            &0.9653        & \bf{0.9663} \\
 \bottomrule
  \end{tabular}
\vspace{-2mm}
\end{table}

\section{Conclusion}
%
Motivated by the convolution theorem, we have presented an effective and efficient method that explores the properties of Transformers for high-quality image deblurring.
We have developed an efficient frequency domain-based self-attention solver (FSAS) to estimate the scaled dot-product attention by an element-wise product operation instead of the matrix multiplication in the spatial domain, where we show that the spatial complexity and the computational complexity are significantly reduced.
We further propose a DFFN to discriminatively determine which low and high frequency information of the features should be preserved for latent clear image restoration.
Moreover, we develop an asymmetrical network based on an encoder and decoder architecture, where the FSAS is only used in the decoder module for better image deblurring.
By training the proposed method in an end-to-end manner, we show that it performs favorably against the state-of-the-art approaches in terms of accuracy and efficiency.

{\small
\bibliographystyle{ieee_fullname}
\bibliography{egbib}
}

\end{document}